\documentclass[letterpaper]{article} 
\usepackage[preprint]{aaai2027}  
\usepackage[hyphens]{url}  
\usepackage{graphicx} 
\usepackage{natbib}  
\usepackage{caption} 
\usepackage{algorithm}
\usepackage{algorithmic}
\usepackage{subcaption}
\usepackage{amsmath}
\usepackage{tikz}
\usepackage{pgfplots}
\usepackage{multirow}
\usepackage{amssymb}
\usepackage{array}
\usepackage[T1]{fontenc}
\usepackage[switch]{lineno} 

\usepackage{newfloat}
\usepackage{listings}
\DeclareCaptionStyle{ruled}{labelfont=normalfont,labelsep=colon,strut=off} 
\floatstyle{ruled}
\newfloat{listing}{tb}{lst}{}
\floatname{listing}{Listing}

\usepackage{booktabs}
\newcommand{\avr}{\mathcal{A}}
\title{MSTypography: Multi-character Semantic Typography \\via Balancing Word Legibility and Object Recognizability}
\author {
    Xinye Yang\textsuperscript{\rm 1},
    Xinding Zhu\textsuperscript{\rm 1},
    Kai Fang\textsuperscript{\rm 1},
    Xinyi Ren\textsuperscript{\rm 1},
    Mengjian Li\textsuperscript{\rm 2},
    Bin Cao\textsuperscript{\rm 1},
    Jiazhou Chen\textsuperscript{\rm 1}\corresponding
}
\affiliations {
    \textsuperscript{\rm 1}Zhejiang University of Technology\\
    \textsuperscript{\rm 2}Zhejiang Lab\\
    Email: cjz@zjut.edu.cn
}

\begin{document}
\maketitle

\begin{abstract}
\label{abs}


Semantic typography is a design technique where the visual representation of a word conveys its semantic meaning, while maintaining its legibility. Existing digital typography methods mainly focus on single-character scenarios. They suffer from a lack of legibility constraints and insufficient local deformation when extended to multi-character words, as the intricate structures among multiple characters are hardly preserved during the typography process. In this paper, we propose a global-to-local typography framework for multi-character scenarios. It performs mask-driven silhouette approximation at the global level, while semantic-guided refinement at the local level, with a culling step in between to improve efficiency. To preserve word legibility, we designed structural losses (including explicit collision constraints and implicit Jacobian singular value constraints) and an OCR constraint for character-level readability. To enhance the object recognizability, we leverage semantic guidance with diffusion priors, which drives the character glyph toward the target concept while preserving its structural integrity. To the best of our knowledge, this is the first multi-character semantic typography method that effectively balances word legibility and object recognizability. Evaluations on five representative languages (English, Chinese, Japanese, Korean, Arabic) demonstrate superiority over SOTA methods. Codes will be open-sourced.

\end{abstract}


\section{Introduction}
\label{sec:intro}

\begin{figure}[tbp]
    \centering
    \includegraphics[width=\columnwidth, keepaspectratio]{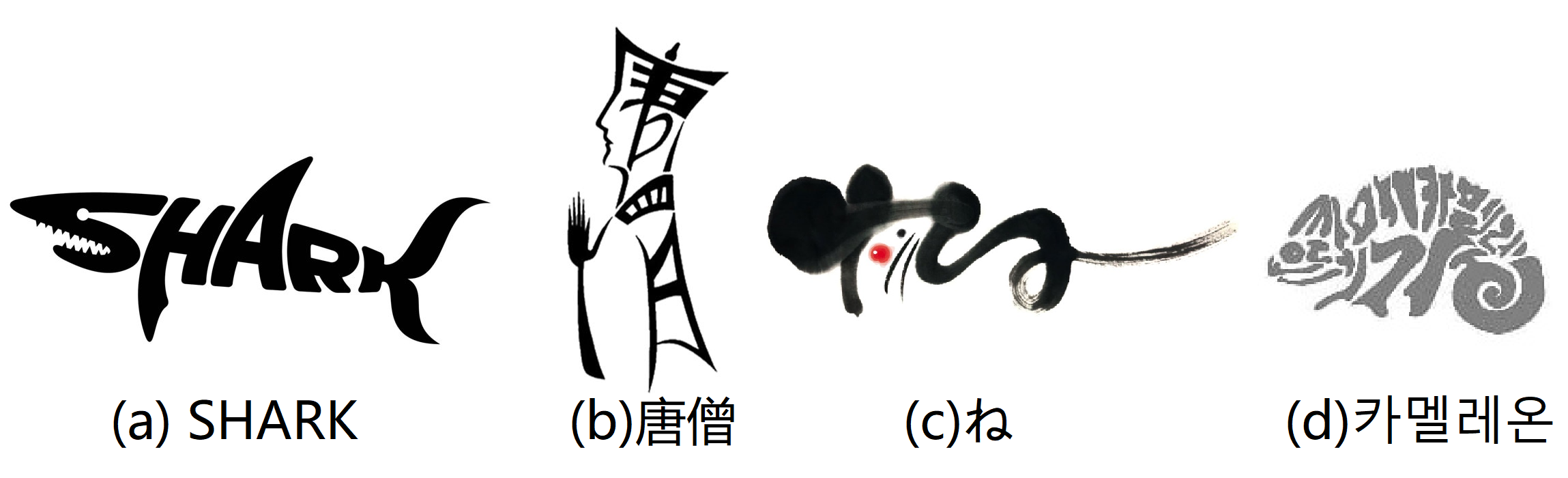}
    
    \caption{The artists' work, (a) English~\cite{sharkenergy_logo}, (b) Chinese~\cite{zhurencen2016art}, (c) Japanese~\cite{hudejii2020japanesecalligraphy} and (d) Korean~\cite{kyriazi2021hangeulart}.}
    \label{fig:art}
\end{figure}

Semantic typography is a design practice that shapes the visual form of a word to reflect its underlying meaning, while still keeping the text readable. For example, turning the letter ``M'' in the word "Mountain" into a mountain, or the Chinese character for ``water'' into a wave. As shown in Fig.~\ref{fig:art}, artists have created such typographic illustrations in different languages (English, Chinese, Japanese, and Korean). This technique has found widespread applications in logo design, book illustration, motion graphics, and creative typography. 


Despite its widespread use, creating typographic illustrations remains labor-intensive, requiring skilled artists and extensive manual refinement. Automation is therefore appealing, yet it inevitably faces a fundamental tension: preserving word legibility while achieving object recognizability. This balance becomes especially challenging with multiple characters, where each character must remain legible while collectively forming a coherent shape.

\begin{figure}[t]
    \centering
    \includegraphics[width=\linewidth]{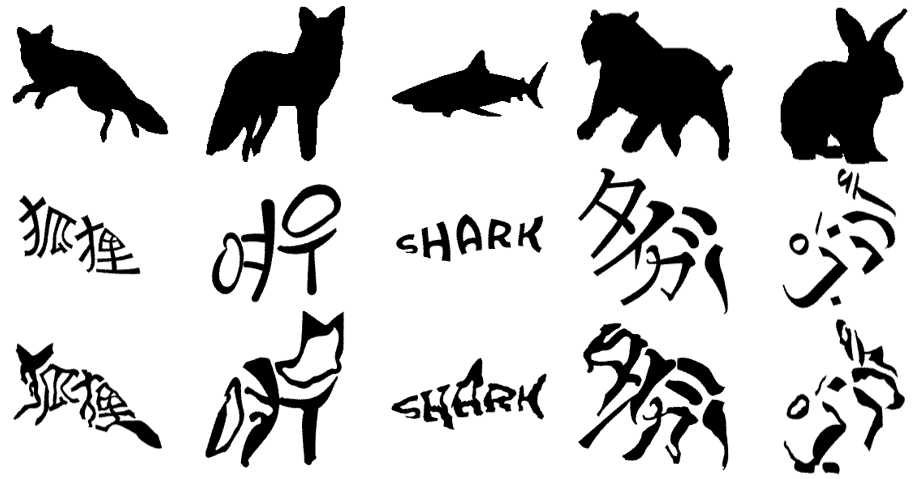}
    \caption{Results of our method with five different languages. From top to bottom are initial masks, intermediate results with global deformation and final typography results. From left to right are foxes in Chinese and Korean, sharks in English, bunnies in Japanese and camels in Arabic.}
    \label{fig:teaser}
\end{figure}

In the last decade, a variety of automatic methods have been proposed for generating word art from text inputs. However, most of them focus primarily on single-character cases. For instance, Word-As-Image~\cite{WordAsImage} and its variant Textured Word-As-Image~\cite{TexturedWordAsImage} optimize letter contours using score distillation sampling (SDS) to produce editable SVGs, while VitaGlyph~\cite{VitaGlyph} introduces a subject environment dual-branch diffusion mechanism. These methods can balance character clarity and semantic fidelity in single-character scenarios. However, when extended to multi-character words, they exhibit three typical artifacts: partial deformation (only some characters are altered), excessive distortion (loss of legibility), and conservative deformation (loss of object recognizability). 
The only attempt for multiple characters is Khattat~\cite{Khattat}. It merely searches locally for low-loss deformation regions rather than treating the whole word as a unified entity, thereby circumventing the core challenge of global semantic typography. Consequently, existing approaches still lack a principled way to produce a coherent, semantically meaningful deformation across all characters while preserving each character's legibility.

In this paper, we propose a global-to-local typography framework that operates at the word level. Fig.~\ref{fig:teaser} shows our results in five representative languages. Our framework begins with an initial layout of a multi-character word, followed by a two-level optimization with a culling step in between. In the global level, a target mask guides the glyphs to stretch freely, overcoming the issue of conservative deformation. The culling step then selects the most promising intermediate results based on their alignment with the mask, discarding poorly deformed instances to improve efficiency and provide high-quality initializations for the next level. In the local level, we introduce ControlNet for global target guidance, whose SDS loss drives the glyphs toward the target semantics, while OCR is employed to enforce character-level readability constraints, preventing excessive distortion and partial deformation. Throughout the optimization, structural losses tailored for closed Bézier curves are embedded, including explicit collision constraints and implicit Jacobian singular value constraints, to maintain glyph topology and morphological stability.

The main contributions of this paper are as follows:


\begin{itemize}


\item \textbf{A global-to-local typography framework for multi-character words}, built on differentiable vector graphics rendering and effectively balancing word legibility and object recognizability across diverse languages.

\item \textbf{Mask-guided layout for semantic approximation}. A target mask is leveraged to guide the initial layout of glyphs and progressively refines their arrangement to achieve coherent semantic alignment as a whole.

\item \textbf{Local structural constraints for legibility preservation}. We integrate OCR supervision with explicit geometric losses and implicit Jacobian constraints to maintain glyph integrity throughout optimization.
\end{itemize}


\section{Related Works}
\label{sec:realted}

\subsection{Semantic Typography}
\label{subsec:sem}

In recent years, several studies have advanced the field of semantic typography. Word-As-Image~\cite{WordAsImage} leverages diffusion priors to optimize Bézier curves, enabling glyphs to visually convey their meaning, but it optimizes a single word as a whole, lacking independent control over individual characters and spatial coordination. Khattat~\cite{Khattat} extends this idea to multi‑character scenarios, achieving end‑to‑end stylization across multiple languages via large language models and an OCR loss, yet its reliance on the OCR penalty restricts the degrees of freedom of the Bézier curves, resulting in conservative deformation magnitude and insufficient semantic expression.

In addition, numerous studies has been explored for glyph stylization, special effects, animation, and domain specific applications. In the area of style generation and font design, FontCrafter~\cite{FontCrafter} proposes an element‑based framework; ArtGlyphDiffuser~\cite{ArtGlyphDiffuser}, FontStudio~\cite{FontStudio}, VitaGlyph~\cite{VitaGlyph}, and UniCalli~\cite{Unicalli} employ cross‑modal fusion, shape‑adaptive diffusion, dual branch diffusion, and a unified framework, respectively, to achieve high‑quality artistic typography, effect rendering, and handwriting style customization. For animations, Dynamic Typography~\cite{DynamicTypo} adds dynamic effects to text; TypeDance~\cite{TypeDance} focuses on logo design and OBI-Designer~\cite{OBI} is dedicated to stylize oracle bone inscription.
However, most of them focus on single characters and do not explicitly address spatial layout and coordinated deformation among multiple characters.

However, aforementioned progresses did not solve the limitation for the extension of multi character phrases.
Although Khattat supports multiple characters, its conservative constraints (such as the OCR loss) restrict the degrees of freedom for deformation; and existing methods generally lack geometric constraints, leading to self‑intersection, collapse, or loss of legibility. Therefore, a new method that explicitly combines spatial arrangement with geometric constraints is needed to construct a unified differentiable framework that optimizes both global layout and local glyph deformation.

\subsection{Vector Graphics Generation}
\label{subsec:vg}

Diffusion-based text-to-image models, when combined with spatial condition maps, enable structure-guided generation. ControlNet~\cite{ControlNet} and its extensions~\cite{T2IAdapter, ControlNetXS, FreeControl, DCControlNet, FineControlNet, DivControl} provide effective ways to inject edge maps, depth maps, or segmentation masks into the generation process. Complementary works~\cite{VODiff, Storm, SIGMA, OmniGen, OminiControl} also improve layout fidelity by modeling visibility order or multi-condition interactions. However, all these methods operate on pixel representations. Pixel-based generation lacks explicit constraints on geometric properties such as stroke width, topological integrity, and character-specific structure. Consequently, directly applying them to multi glyph semantic typography often leads to inter glyph interference, stroke collapse, or loss of legibility, nor can they directly produce editable vector graphics.

\begin{figure*}[htbp]
    \centering
    \includegraphics[width=\linewidth]{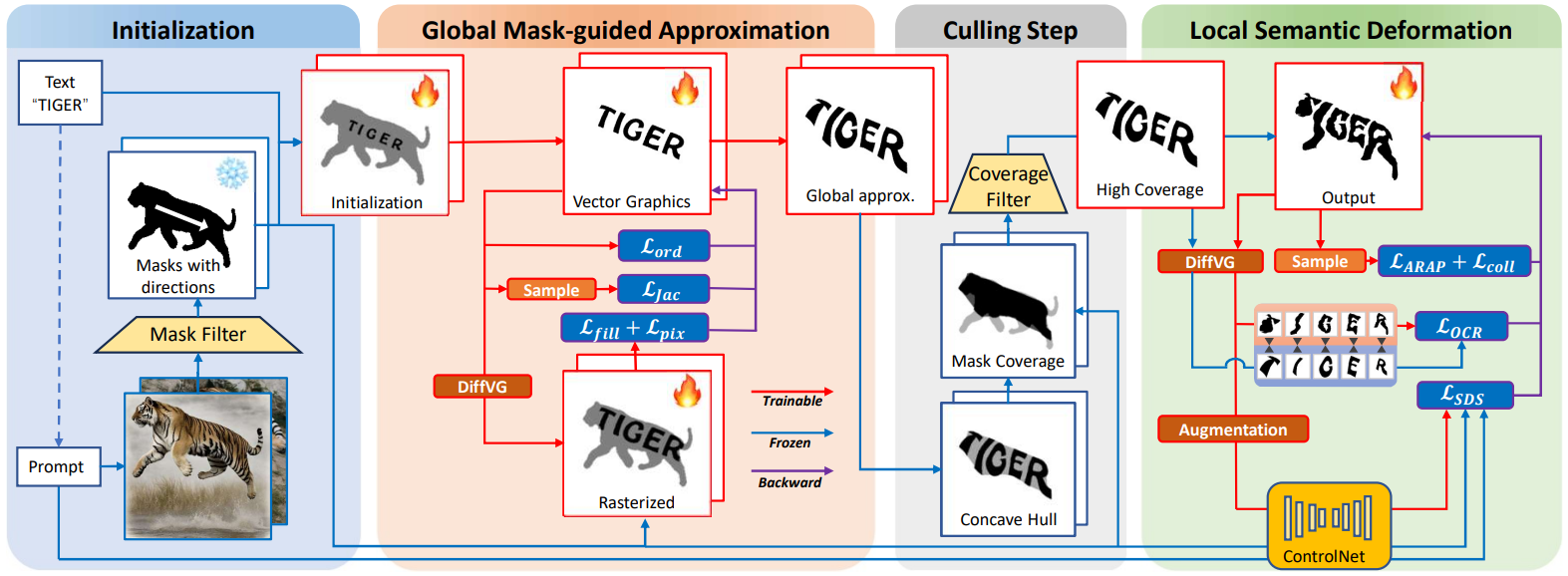}
    \caption{The overview of our global-to-local semantic typography framework. At the global level, PCA-based initial placement, linear/Bézier deformations optimized by mask-filling, layout, and Jacobian losses. A culling step then filters candidates by concave-hull IoU. At the local level, semantic guidance for detail refinement, with OCR, collision detection, and ARAP constraints to preserve legibility and local rigidity.}
    \label{fig:overview}
\end{figure*}

As an alternative, vector graphics generation based on differentiable rendering is an important technical route. This approach builds on DiffVG~\cite{DiffVG} to rasterize Bézier curves into images and iteratively optimizes graphical parameters using vision‑language model losses or the SDS loss. From CLIPDraw~\cite{CLIPDraw} to VectorFusion~\cite{VectorFusion} and SVGDreamer~\cite{SVGDreamer}, this paradigm has progressively improved semantic consistency and generation diversity. However, these early methods primarily target single glyph or simple shapes, lacking explicit constraints on spatial coordination among multiple glyphs and the ability to preserve the legibility of each glyph independently.

To mitigate the problem of shape decomposition caused by numerous overlapping paths during optimization, NeuralSVG~\cite{NeuralSVG} and SVGDreamer++~\cite{SVGDreamer++} introduce implicit regularization and adaptive primitive count, respectively. DuetSVG~\cite{DuetSVG} further proposes a dual branches collaborative generation framework to enhance topological quality and editability. Nevertheless, these methods still cannot actively avoid inter glyph overlaps nor maintain the legibility of individual characters during deformation. In this paper, we propose a global‑to‑local optimization, active layout, and geometric constraints to specifically address the balance problem in multi‑character semantic typography.

\section{Methodology}
\label{sec:method}


Our method follows a global-to-local optimization paradigm, as illustrated in Fig.~\ref{fig:overview}. In the global level, we perform overall morphological adjustments on the input vector glyphs through linear transformations and nonlinear Bézier deformations, with the common goal of making the glyphs globally approximate the target object image as closely as possible. In the local level, we employ ControlNet to guide Stable Diffusion for image generation, while introducing OCR and morphological constraints to preserve glyph structures.


\subsection{Initialization}
\label{subsec:init}


In the initialization, we first generate an image reflecting the shape of the target object from a semantic text prompt using a diffusion model and then extract its main region via image segmentation as the input mask image $\mathbf{M}$. We then convert the input characters into B\'ezier curves and carry out pre-layout operations: we compute the principal orientation of the main body (i.e., the black region) of $\mathbf{M}$ (of size $C \times H \times W$) using PCA~\cite{pearson1901lines}, arrange the characters along this orientation, and scale them to avoid collisions. Subsequently, we generate a mesh from the sampled points of the original glyph $\mathbf{G}$ using Delaunay triangulation~\cite{lee1980two}, where each B\'ezier curve is discretized into $m$ samples. For $i$-th glyph $\mathbf{G}_i$, its outline is discretized into $N_i$ samples, defined as $\mathbf{P}_i = [\mathbf{p}_{i,1}, \mathbf{p}_{i,2}, \dots, \mathbf{p}_{i,L_i}]$. These samples are then divided into $C_i$ line segments, denoted as $\mathbf{q}_{i, 1}, \mathbf{q}_{i, 2}, ... \mathbf{q}_{i, C_i}$, where each segment $\mathbf{q}_{i, j}$ connects two consecutive samples $(\mathbf{p}_{i, j}, \mathbf{p}_{i, j+1})$. For closed contours, the last segment connects $\mathbf{p}_{i, N_i}$ back to $\mathbf{q}_{i, 1}$.



\subsection{Global Mask-guided Approximation}
\label{subsec:global}

This level performs initial typography and deformation on the input glyphs to make their overall shape conform to the target silhouette, providing a good starting point for subsequent local optimization. The raw glyphs often deviate significantly from the target shape, and adjusting only their own control points makes it difficult to achieve large pose changes. Therefore, this level combines linear and nonlinear transformations: optimizing translation and scaling for rigid global alignment, while introducing a B\'ezier grid with bilinear interpolation and Coons correction~\cite{Smooth, Curves} to apply smooth nonlinear deformation. The optimization is primarily driven by the fill coverage of the glyphs with respect to the target mask, embedding the glyphs into the rough framework of the target object. The total loss of the global level is:

\begin{align}
\mathcal{L}_{\text{glob}} = \mathcal{L}_{\text{fill}} + \lambda_{\text{ord}}\mathcal{L}_{\text{ord}} + \lambda_{\text{pix}}\mathcal{L}_{\text{pix}} + \lambda_{\text{Jac}}\mathcal{L}_{\text{Jac}}.
\end{align}

For longer words (e.g., in English), we bind adjacent glyphs into joint optimization groups to reduce computational complexity and improve optimization efficiency.

\subsubsection{Object recognizability.} A common approach for semantic guidance in text-to-image generation is Score Distillation Sampling (SDS). However, it suffers from stochastic gradient noise during optimization and is often unstable in providing semantic guidance for multi-character scenarios, causing severe fluctuations of transformation parameters and difficulty in converging to the target shape. To reduce uncertainty, SDS is not adopt in this level; instead, we directly use mask approximation: generating a binary mask from the input image, computing the mean squared error between the deformed rendered image and the mask, and additionally penalizing overflow. The overflow $\rho$ is calculated:

\begin{align}
\rho = \frac{1}{|\Omega_1|} \sum_{i \in \Omega_1} (1 - X_i),
\end{align}

\noindent where $\Omega_1 = \{ i \mid M_i = 1 \}$, and $X_i$ denotes the $i$-th pixel of the image $\mathbf{X}$. When $\Omega_1 = \varnothing$, $\rho$ is set to $0$. The loss is:

\begin{align}
\mathcal{L}_{\text{fill}}(\mathbf{X}, M) = \operatorname{MSE}(\mathbf{X},M) +  \frac{\rho}{1 - \rho}.
\end{align}

\subsubsection{Word legibility.} To maintain spatial coordination among glyphs, we employ an ordering loss to enforce consistent centroid directions and avoid sharp turns, along with a pixel-based layout loss that combines area uniformity and overlap penalties from rendered glyph masks to balance sizes and separate characters. The losses are:

\begin{align}
\mathcal{L}_{\text{ord}} = \operatorname*{\avr}_{\substack{i}} \left( \left(1 - \mathbf{u}_i \cdot \mathbf{d} \right) ^ 2 \right) + \operatorname*{\avr}_{\substack{i}} \left( \left (1 - \mathbf{u}_i \cdot \mathbf{u}_{i+1} \right) ^ 2 \right),
\end{align}

\begin{align}
\mathcal{L}_{\text{pix}}=\operatorname*{\avr}_{\substack{i}} \left((S_i - \bar{S})^2 \right) + \lambda_{\text{over}} \operatorname*{\avr}_{\substack{i, j}} \left ( \frac{|A_i \cap A_j|}{|A_i \cup A_j|} \right ),
\end{align}

\noindent where $\operatorname{\avr}(\cdot)$ represents the average, $\mathbf{u}_i$ denotes the unit direction vector between the $i$-th and $(i+1)$-th glyphs, $\mathbf{d}$ is the unit reference direction vector. $A_i$ represents the area of the $i$-th glyph, $S_i$ denotes the ratio between the current area of the glyph and its initial area, and $\bar{S}$ is the mean of this ratio over the current iteration.

To ensure uniform scaling of the overall glyph and prevent local scaling distortions, we propose a geometric constraint loss based on the Jacobian matrix to regularize the deformation of each triangular face. Specifically, we construct the Jacobian matrix $\mathbf{J}_i$ from the vertex coordinate differences before and after deformation for each face, and use its singular values $\sigma_{i1}$ and $\sigma_{i2}$ to enforce uniform scaling. However, since singular values lack directional information and cannot detect face flipping, we further introduce the determinant $|\mathbf{J}_i|$ to directly penalize flips for this loss:

\begin{align}
\mathcal{L}_{\text{Jac}} = \operatorname*{\avr}_{\substack{i \in [1, N_{f}]}} \left( (\sigma_{i1} - \sigma_{i2})^2 + \lambda_{\text{flip}} \operatorname{ReLU}^2 (-|\mathbf{J}_i|) \right), \label{eq:Jac}
\end{align}

\noindent where $N_f$ denotes the total number of triangular faces.

Under the constraints imposed by these loss functions, our method is able to achieve satisfactory typography results.

\subsection{Culling Step}
\label{subsec:filter}

Due to the stochastic nature of diffusion-based mask generation, globally deformed glyphs do not always align well with the target masks. We attempted to improve the masks using adaptive control strategies such as SmartControl~\cite{liu2024smartcontrol}, but found their effect limited. Moreover, global generation is significantly faster than the subsequent local optimization, creating a computational asymmetry. To address these issues, we introduce a culling step after the global level. We therefore rapidly synthesize a large pool of global candidates and retain only the most promising ones for the expensive local refinement. Concretely, we compute the concave hull of each deformed glyph, fill its interior to obtain a binary image, and measure the IoU with the target mask. Concave hull is adopted because human perception prioritizes global silhouette and contour over internal details when judging shape conformity. All candidates are then ranked by IoU, and the top-$N$ ($N=15$) are selected as inputs for the local level.

Concave hull computation is non-differentiable and therefore cannot be directly optimized in the global level. During global deformation, we instead approximate the alignment using the differentiable fill loss $\mathcal{L}_{\text{fill}}$, which provides stable gradients for iterative optimization. In the subsequent culling step, we replace this proxy with the more accurate concave‑hull IoU to filter out low‑quality results offline. This complementary design allows us to combine iterative optimization feasibility with precise candidate selection.

\begin{figure*}[t]
\centering

\renewcommand{\arraystretch}{0.1}
\setlength{\extrarowheight}{-20pt}
\setlength{\tabcolsep}{2pt}   
\begin{tabular}{
    >{\centering\arraybackslash}m{0.05\textwidth}   
    *{5}{>{\centering\arraybackslash}m{0.085\textwidth}}  
    | 
    *{5}{>{\centering\arraybackslash}m{0.085\textwidth}}  
}
\toprule
{} & English & Chinese & Japanese & Korean & Arabic & English & Chinese & Japanese & Korean & Arabic \\
\midrule

Input &
  \includegraphics[width=0.085\textwidth]{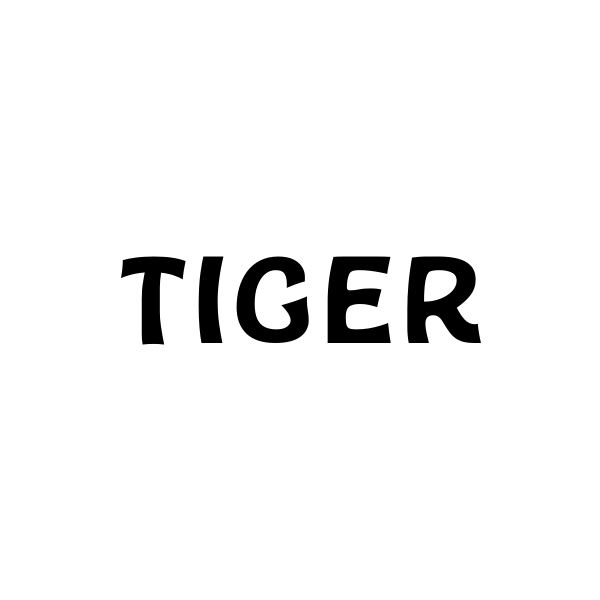} &
  \includegraphics[width=0.085\textwidth]{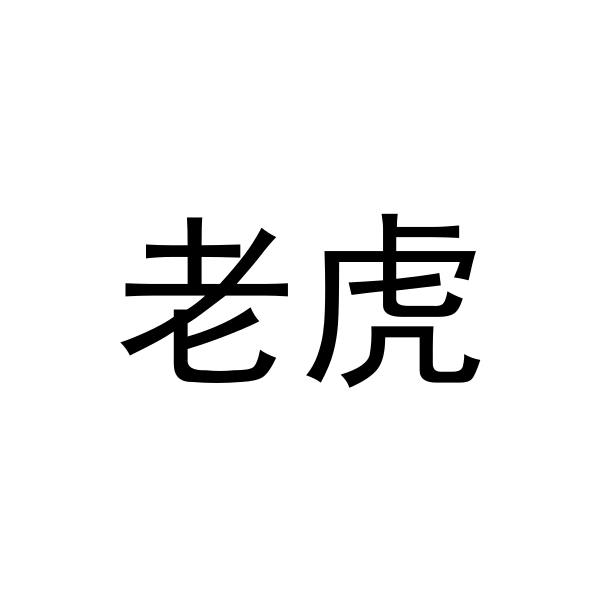} &
  \includegraphics[width=0.085\textwidth]{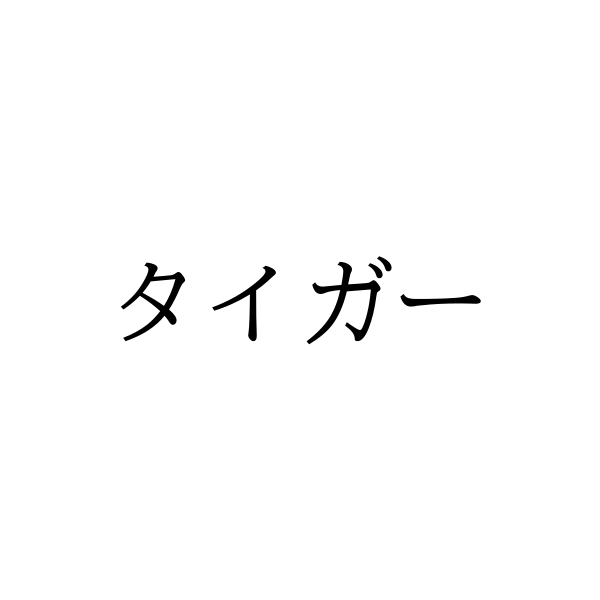} &
  \includegraphics[width=0.085\textwidth]{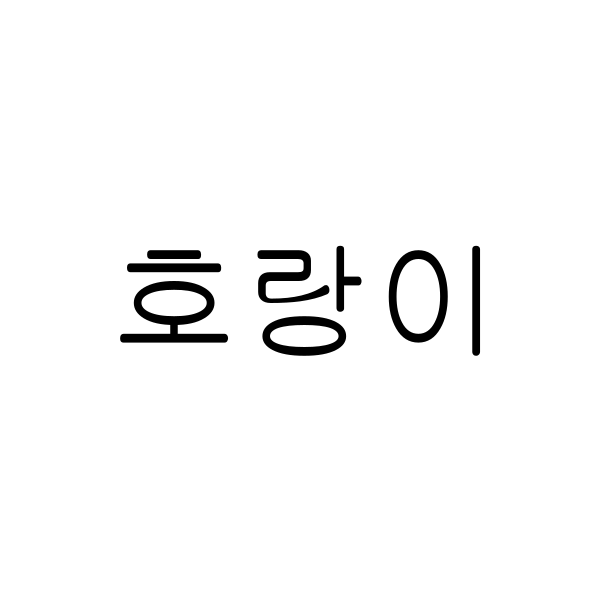} &
  \includegraphics[width=0.085\textwidth]{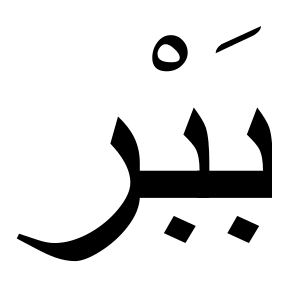} & 
  \includegraphics[width=0.085\textwidth]{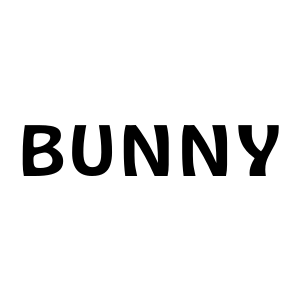} &
  \includegraphics[width=0.085\textwidth]{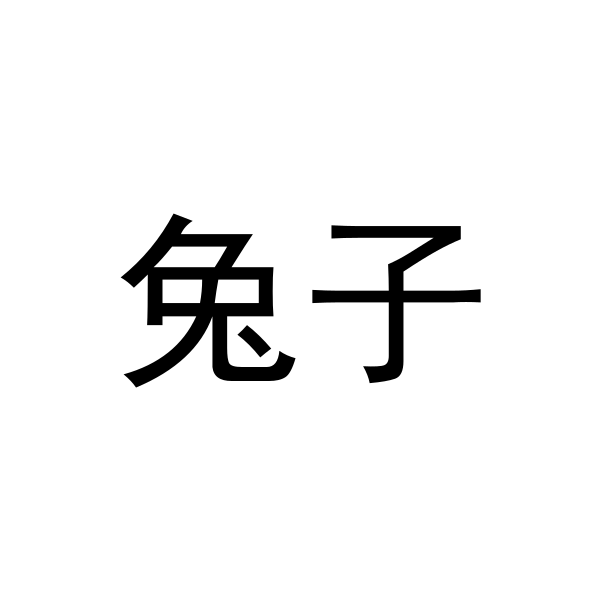} &
  \includegraphics[width=0.085\textwidth]{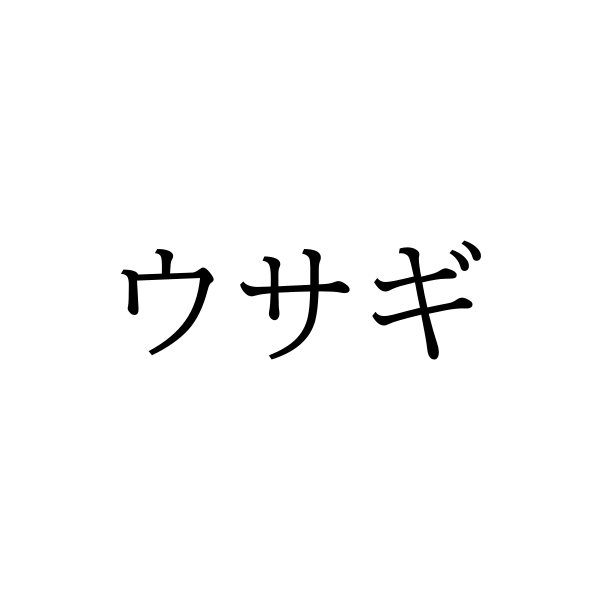} &
  \includegraphics[width=0.085\textwidth]{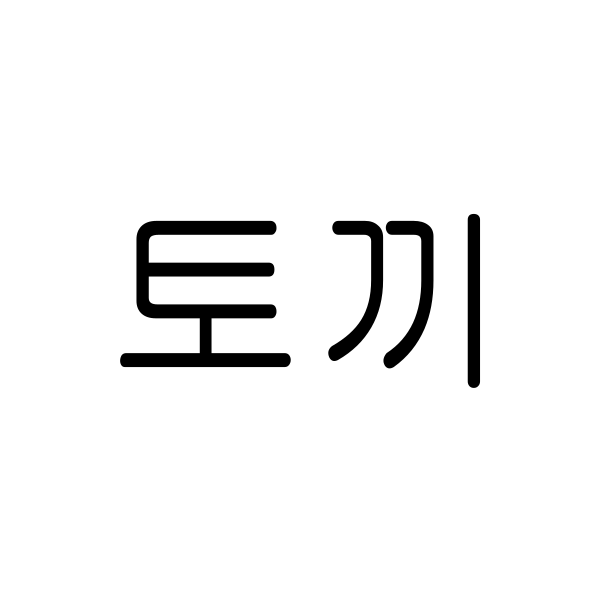} &
  \includegraphics[width=0.085\textwidth]{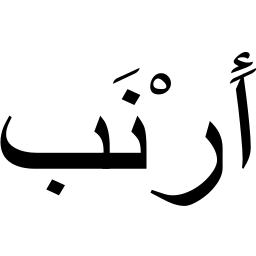} \\

WAI &
  \includegraphics[width=0.085\textwidth]{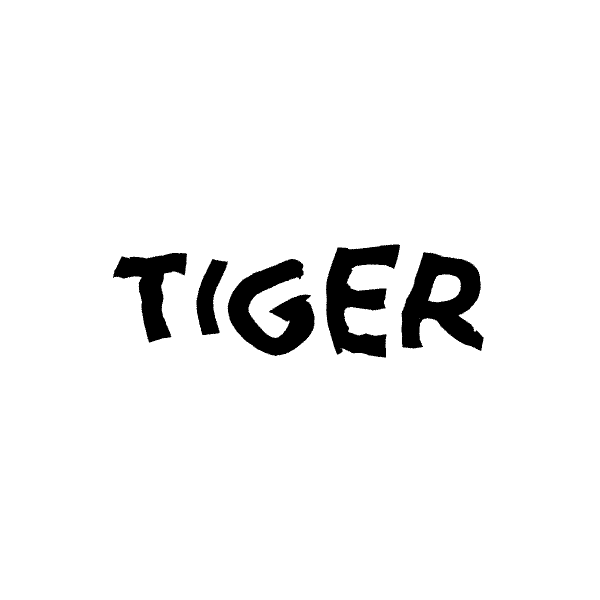} &
  \includegraphics[width=0.085\textwidth]{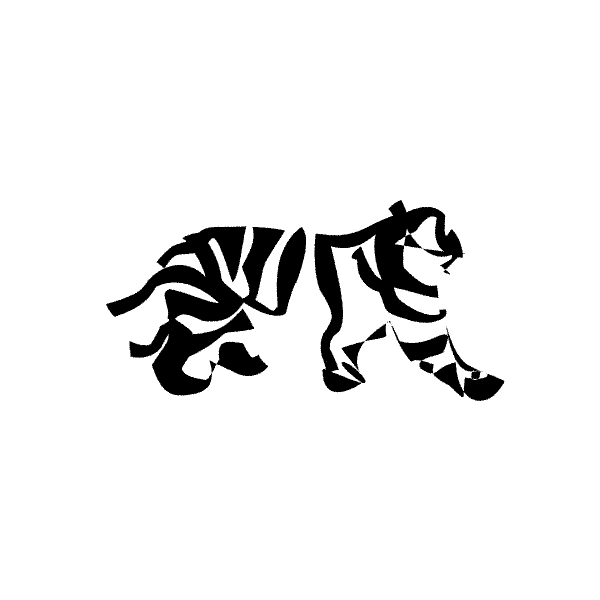} &
  \includegraphics[width=0.085\textwidth]{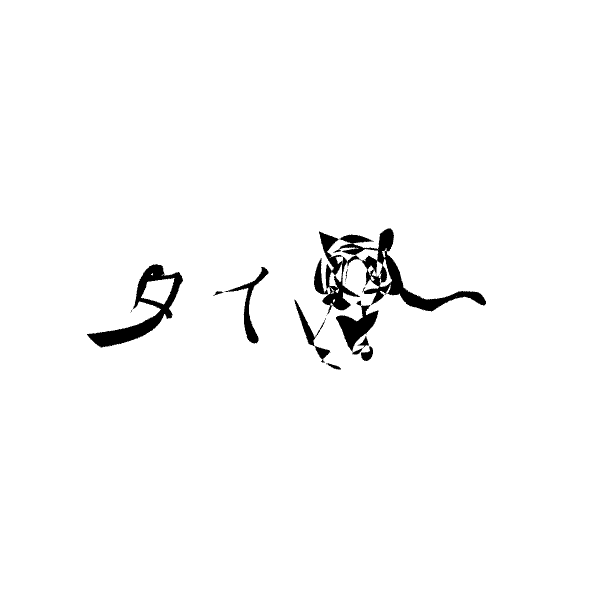} &
  \includegraphics[width=0.085\textwidth]{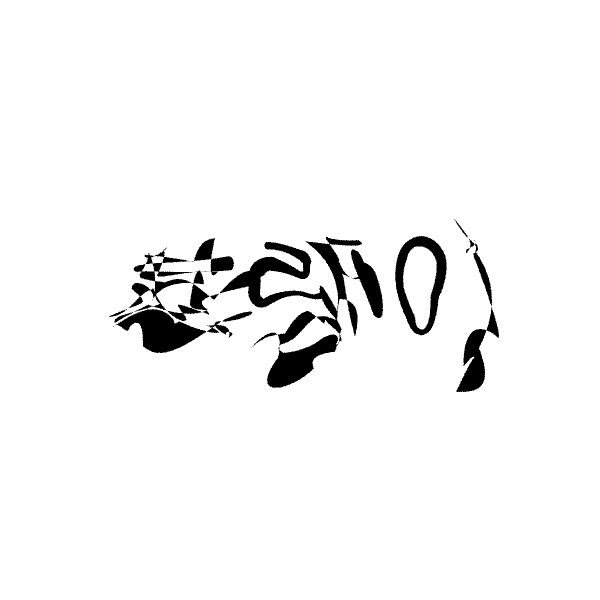} &
  \includegraphics[width=0.085\textwidth]{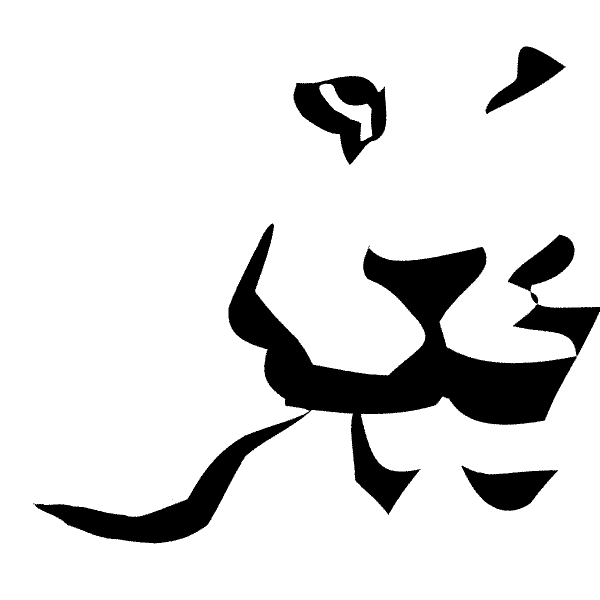} &
  \includegraphics[width=0.085\textwidth]{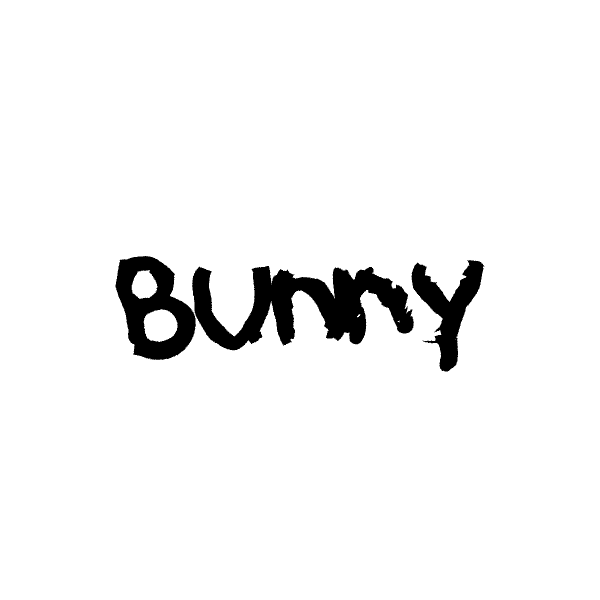} &
  \includegraphics[width=0.085\textwidth]{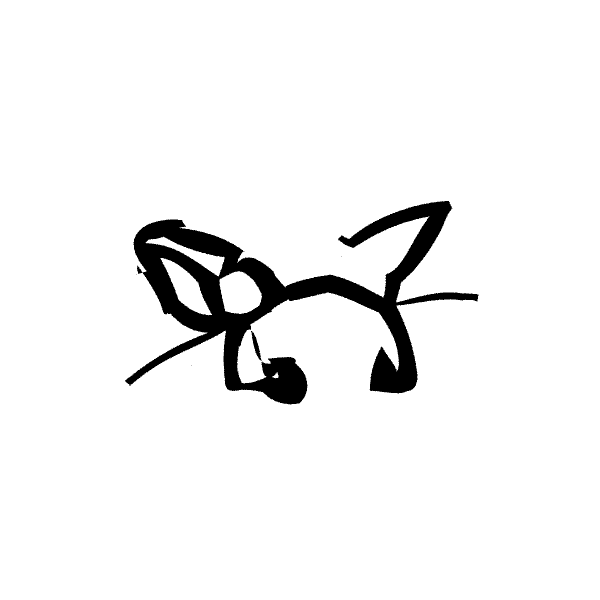} &
  \includegraphics[width=0.085\textwidth]{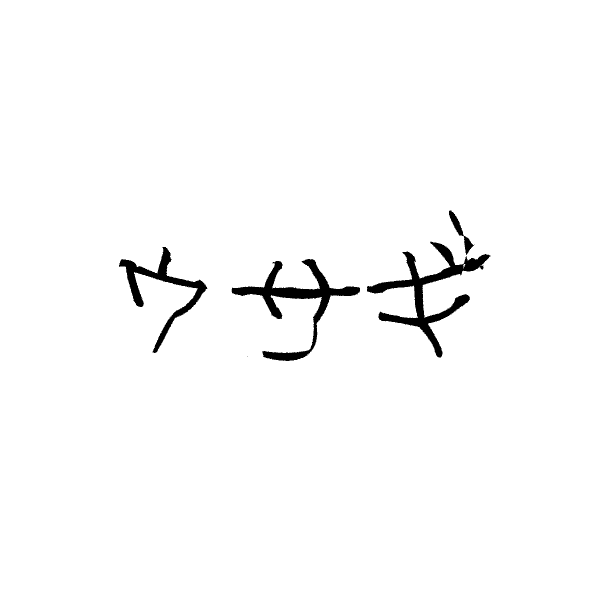} &
  \includegraphics[width=0.085\textwidth]{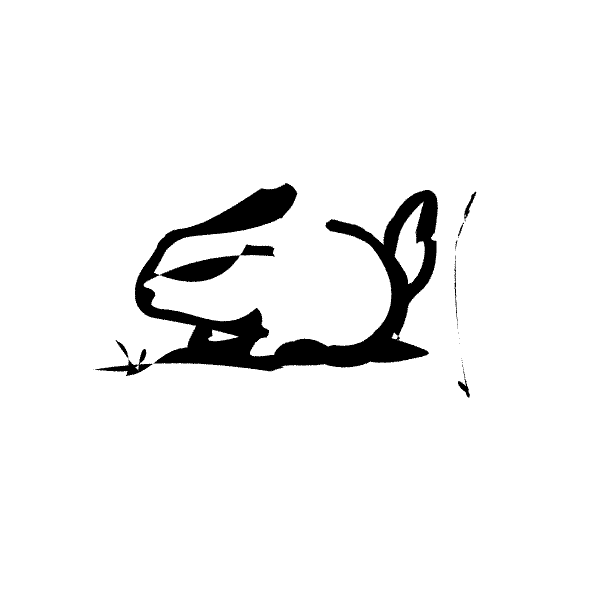} &
  \includegraphics[width=0.085\textwidth]{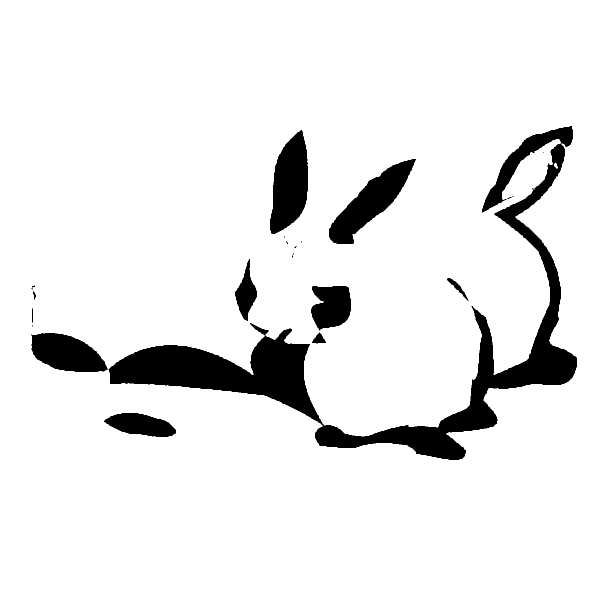} \\

DT &
  \includegraphics[width=0.085\textwidth]{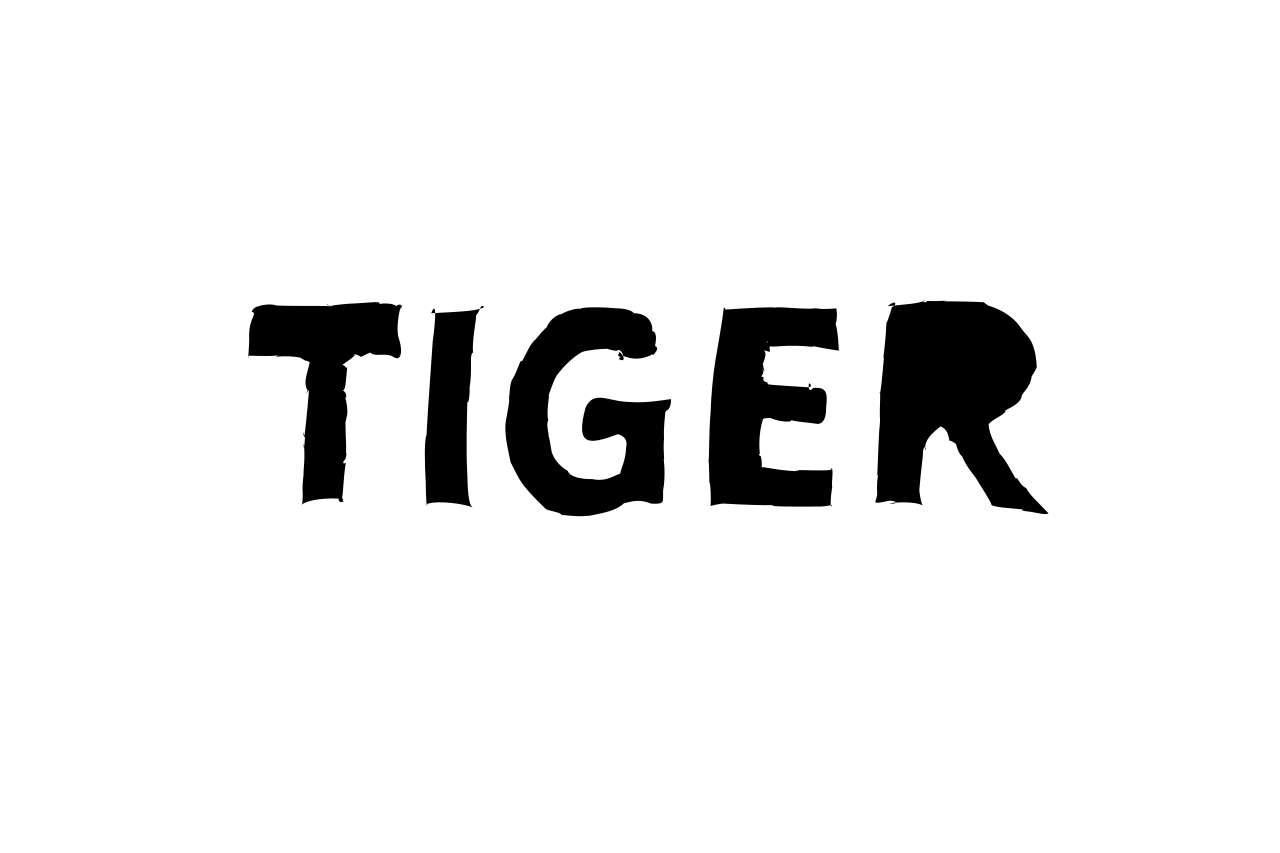} &
  \includegraphics[width=0.085\textwidth]{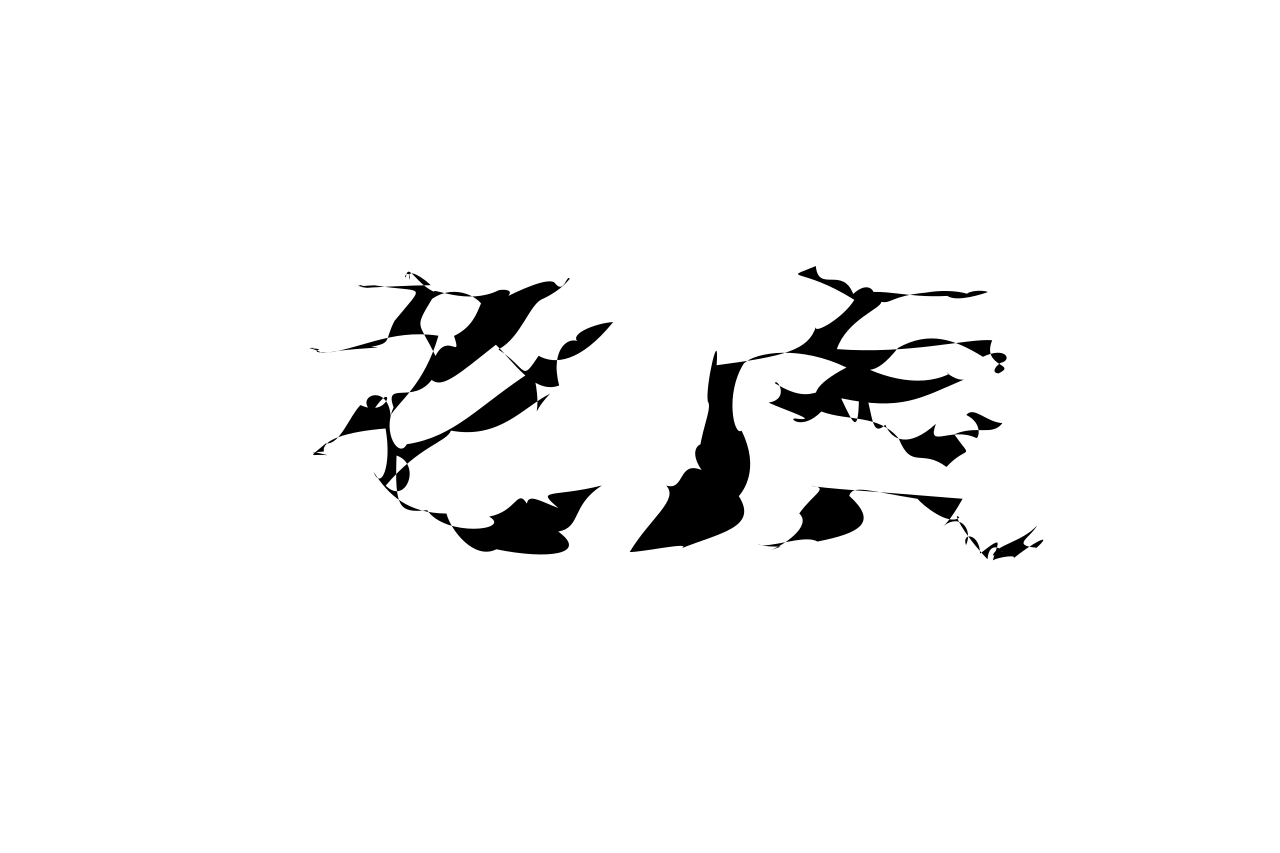} &
  \includegraphics[width=0.085\textwidth]{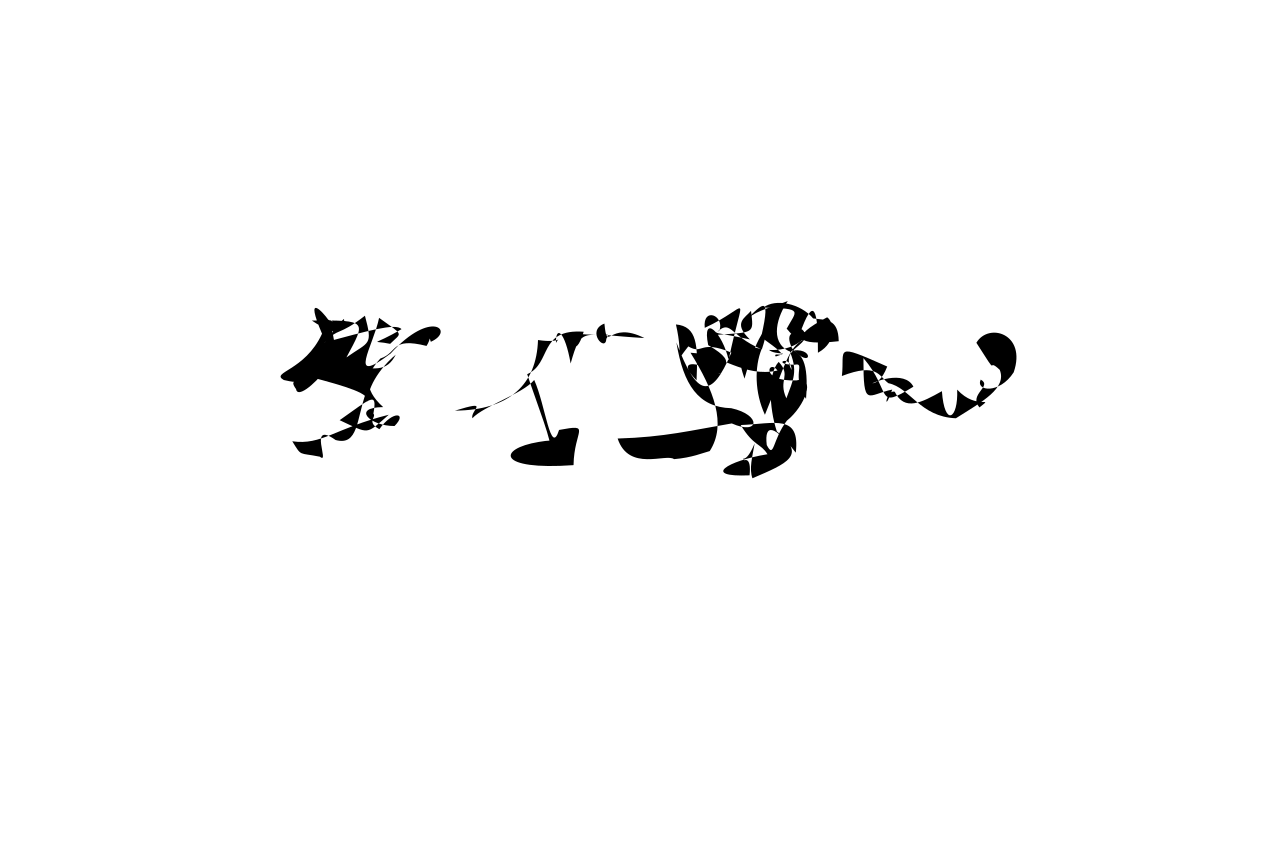} &
  \includegraphics[width=0.085\textwidth]{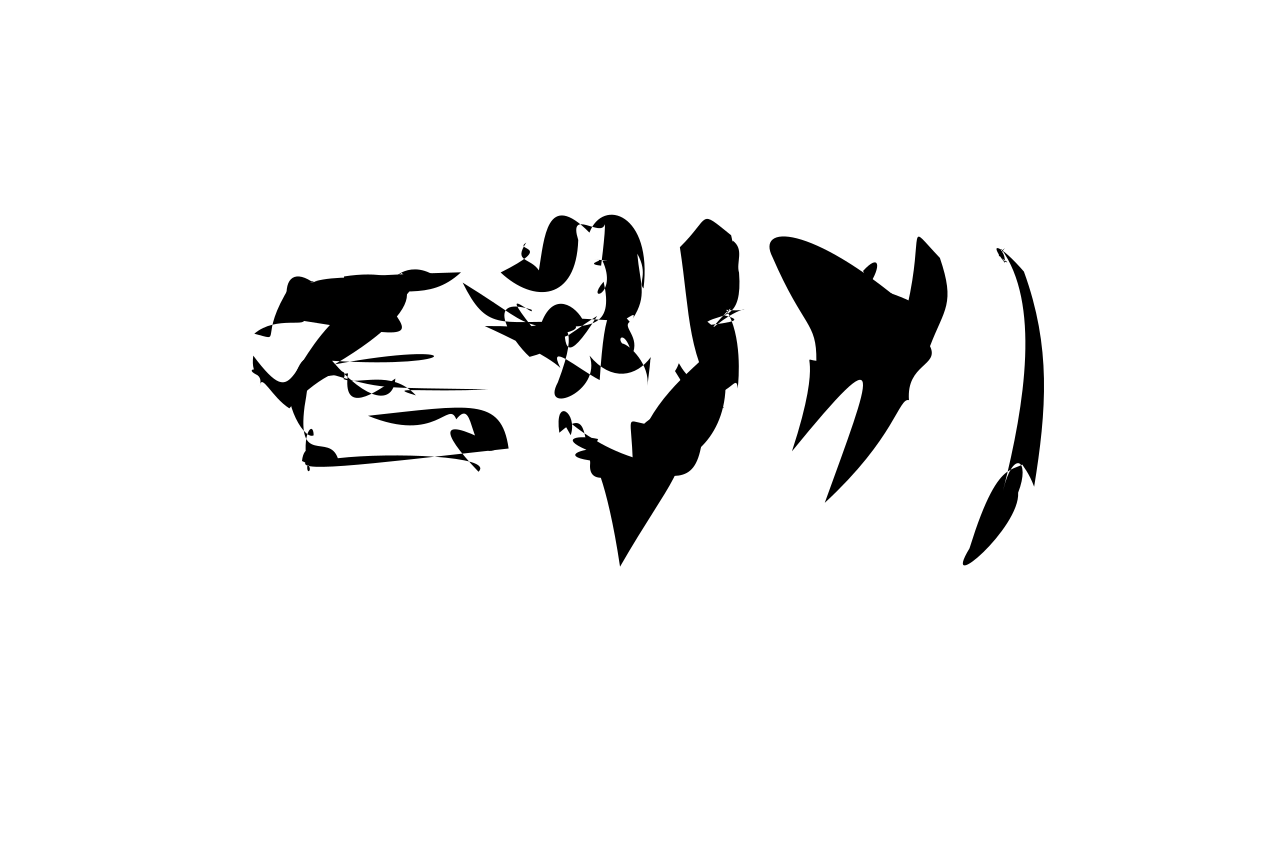} &
  \includegraphics[width=0.085\textwidth]{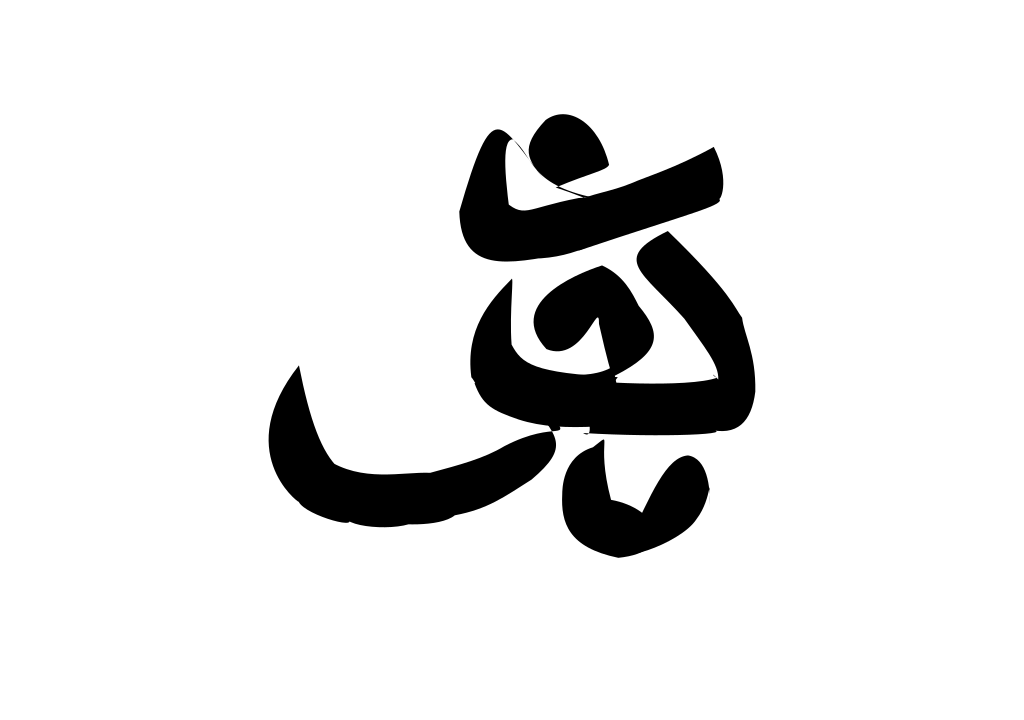} &
  \includegraphics[width=0.085\textwidth]{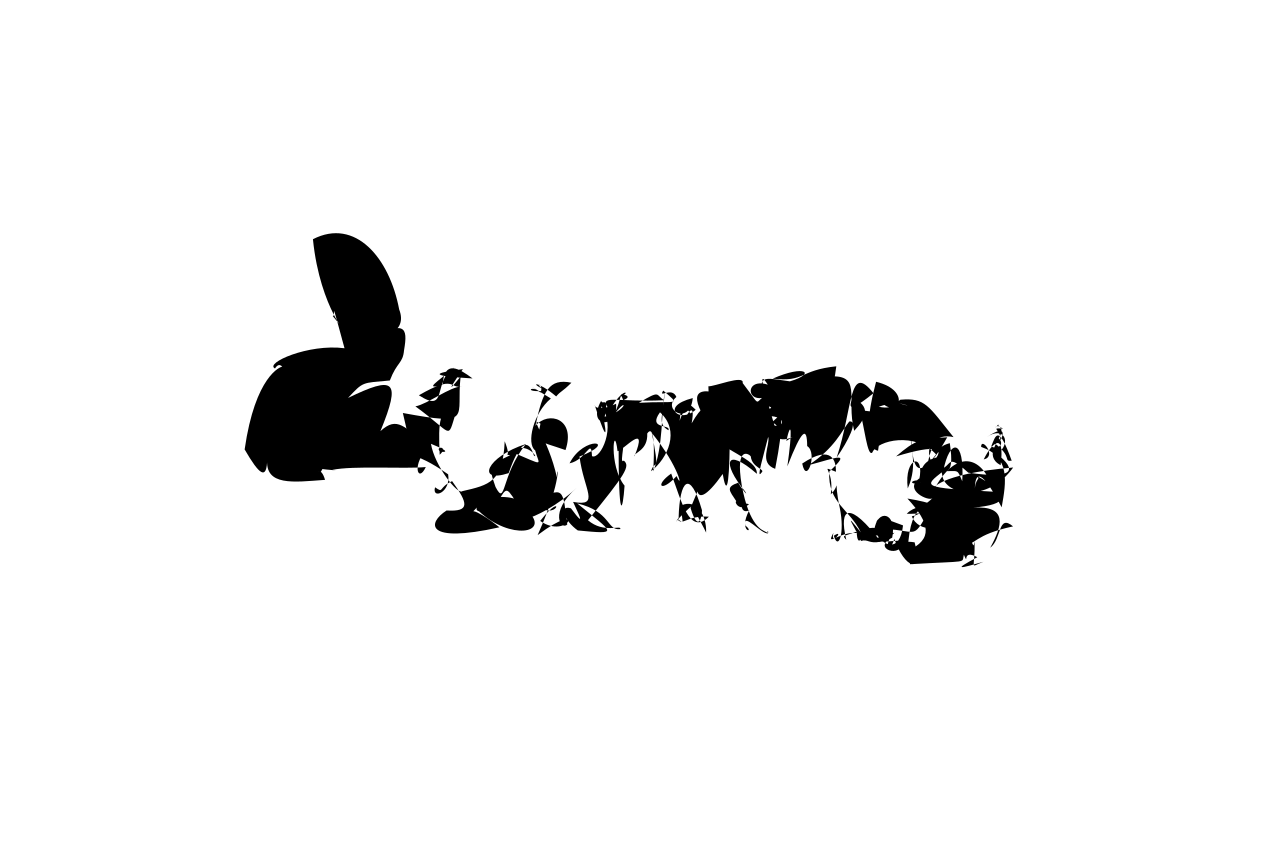} &
  \includegraphics[width=0.085\textwidth]{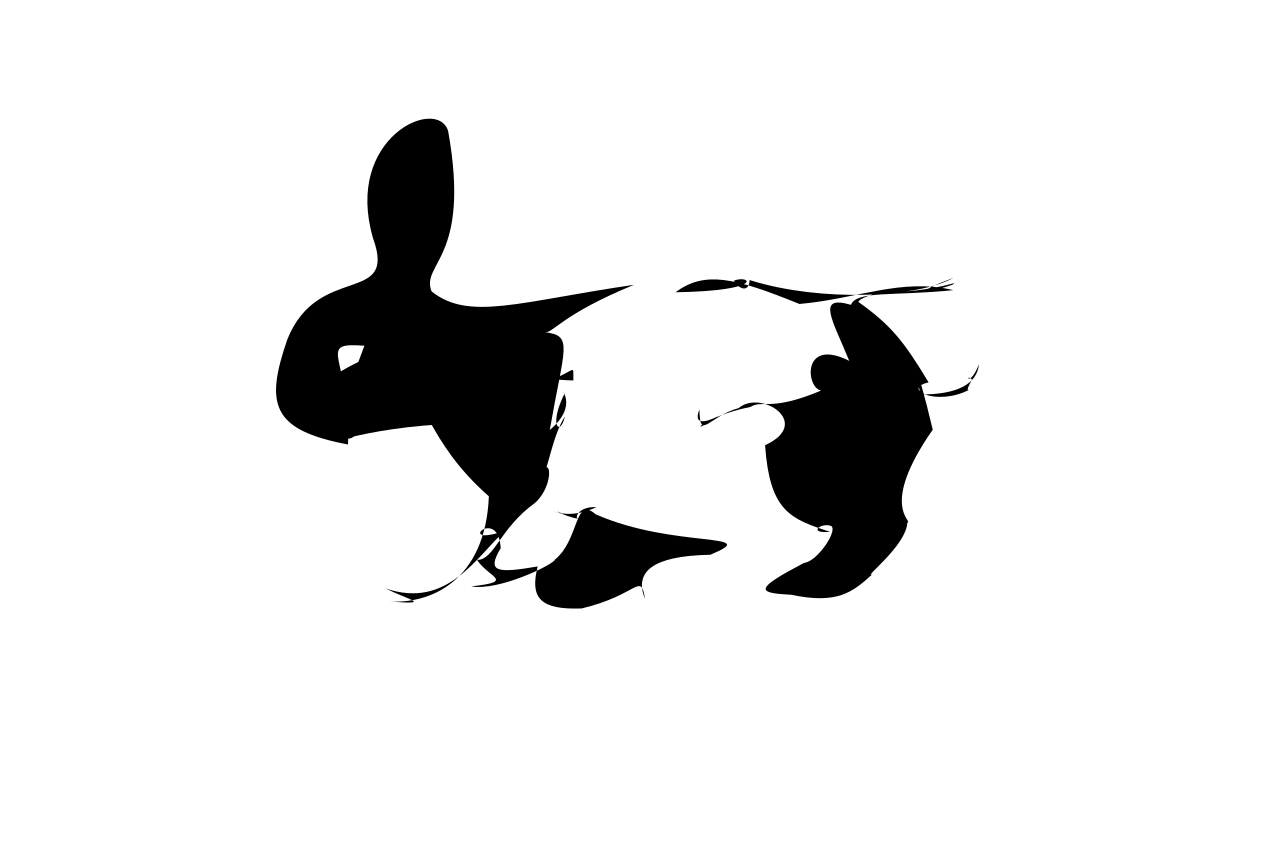} &
  \includegraphics[width=0.085\textwidth]{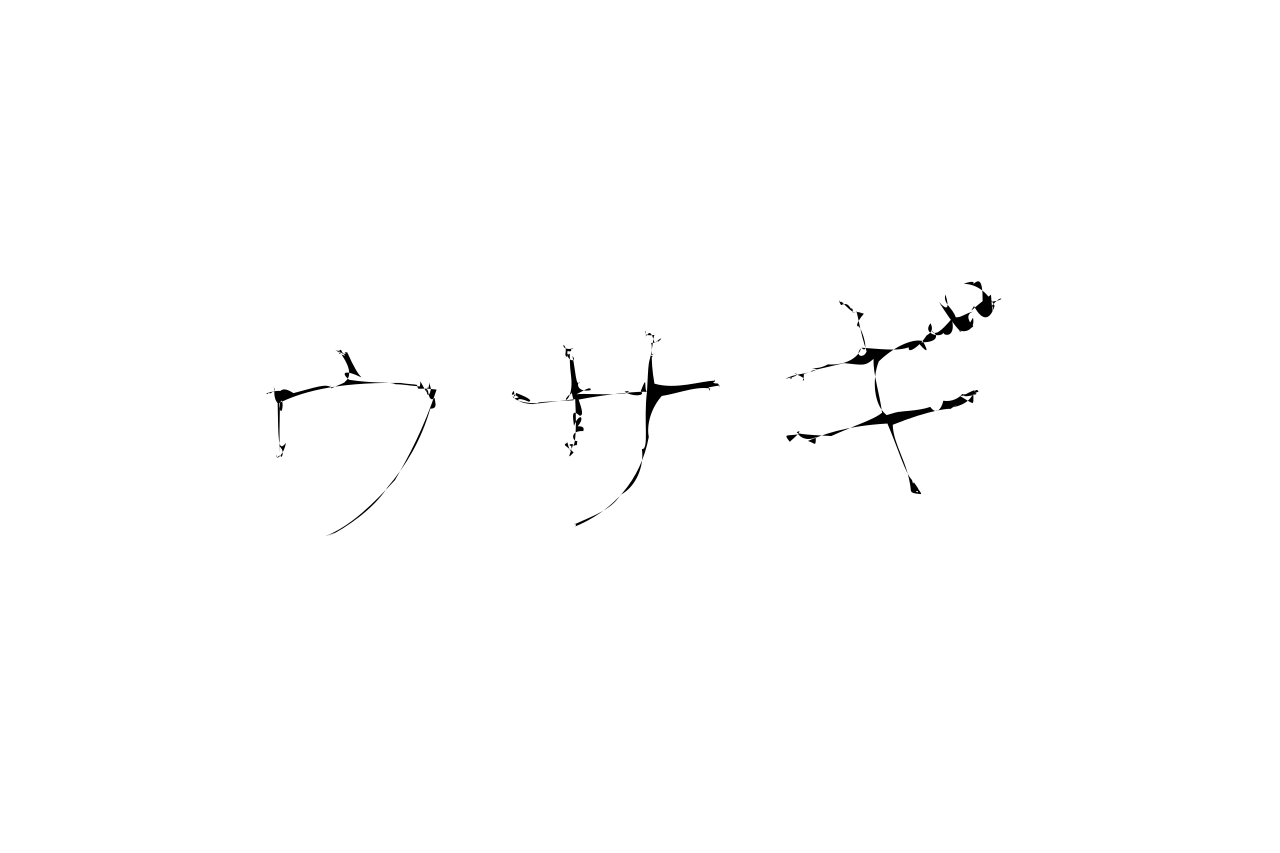} &
  \includegraphics[width=0.085\textwidth]{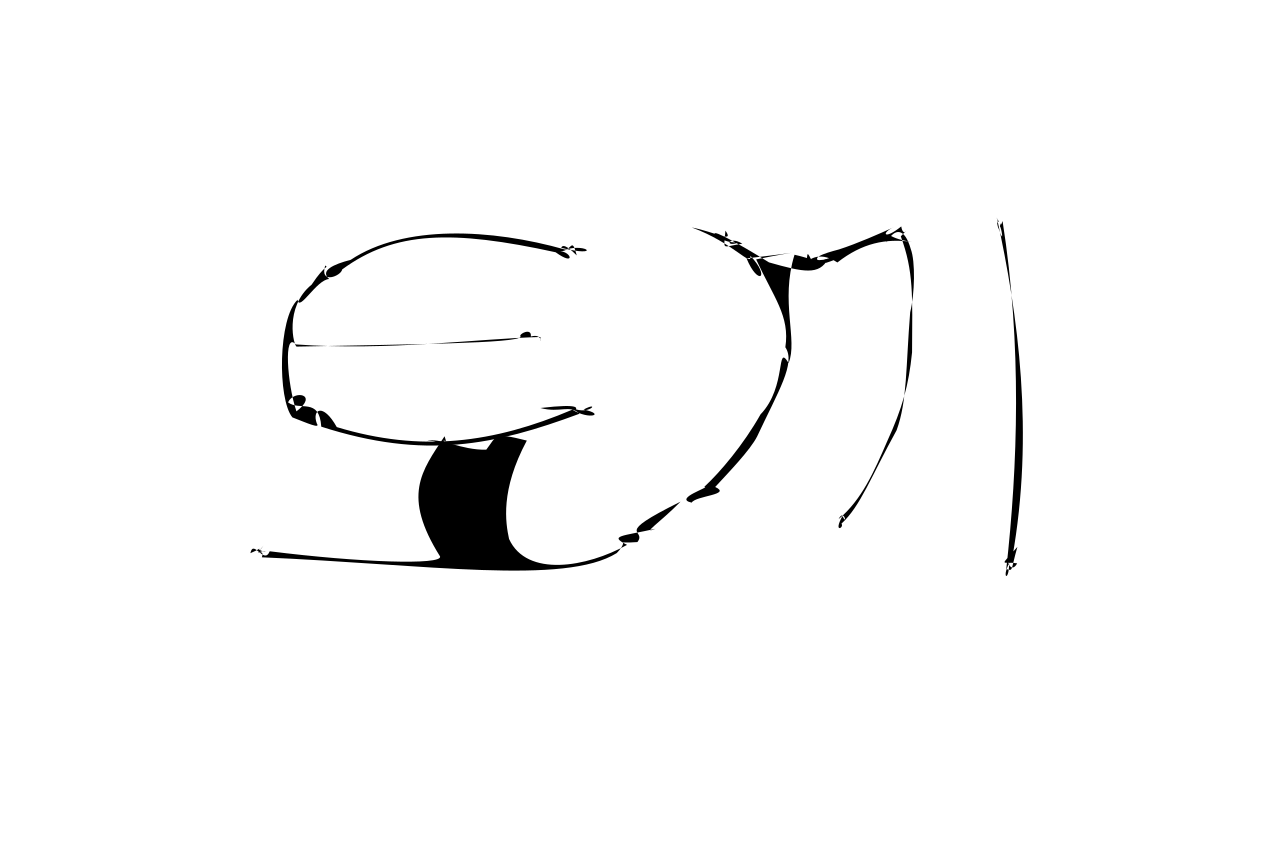} &
  \includegraphics[width=0.085\textwidth]{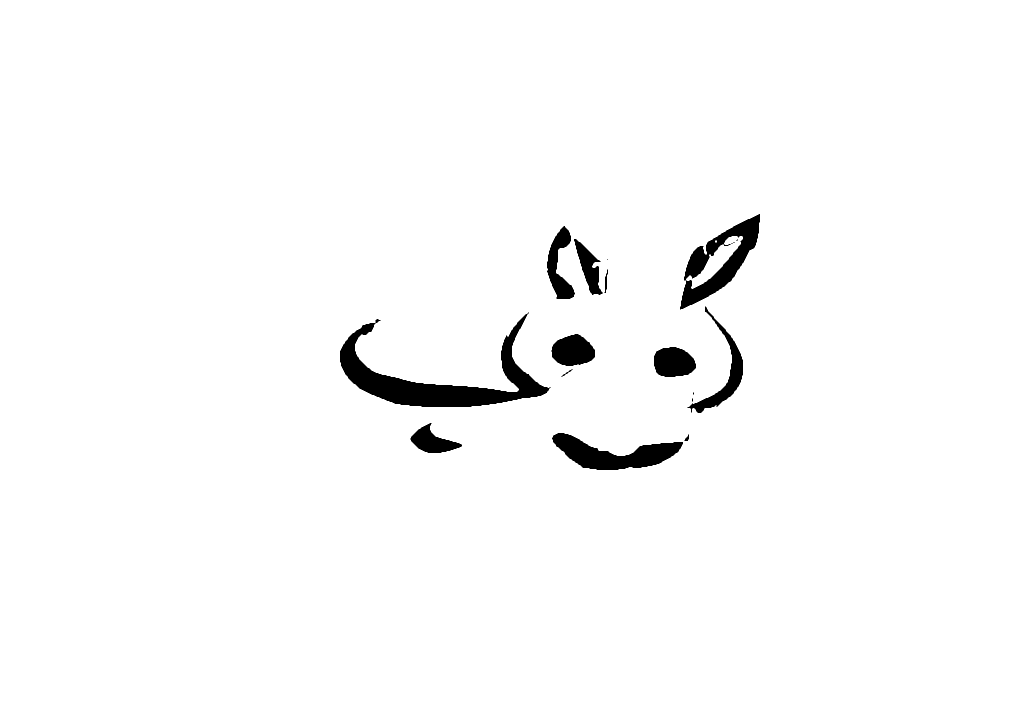} \\

OBI &
  \includegraphics[width=0.085\textwidth]{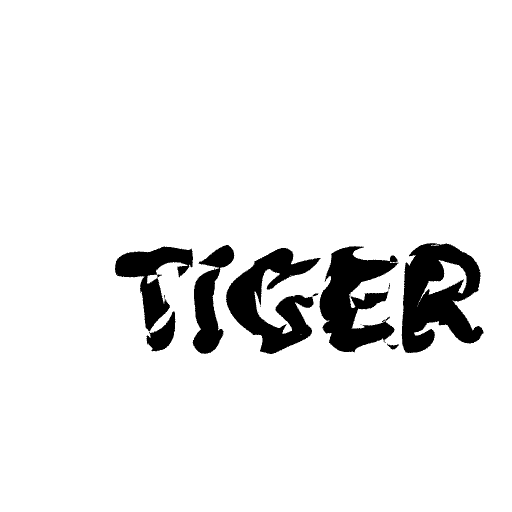} &
  \includegraphics[width=0.085\textwidth]{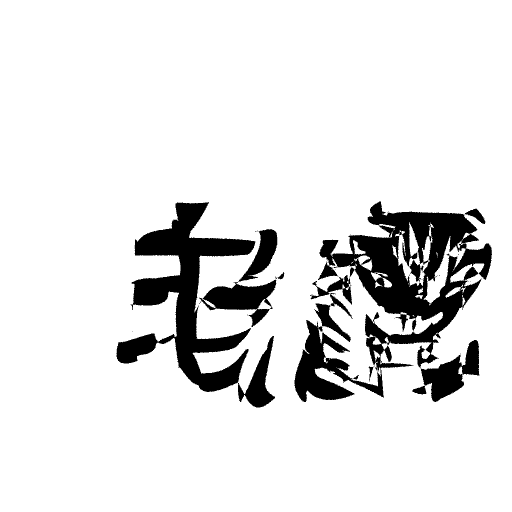} &
  \includegraphics[width=0.085\textwidth]{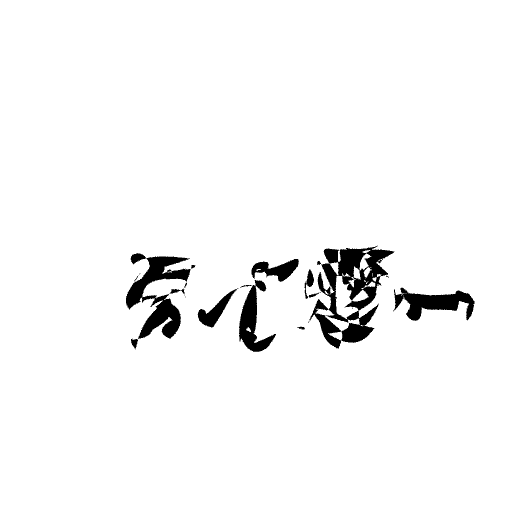} &
  \includegraphics[width=0.085\textwidth]{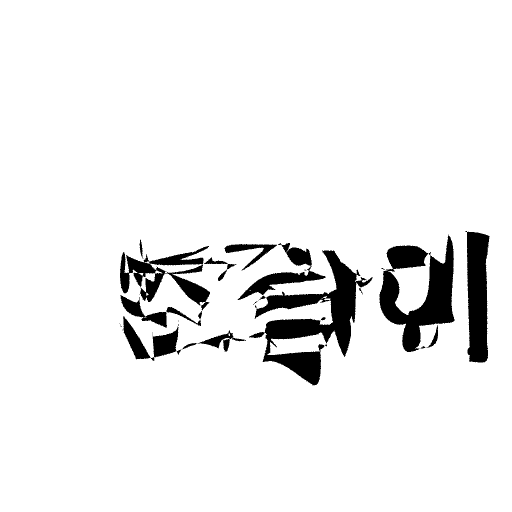} &
  \includegraphics[width=0.085\textwidth]{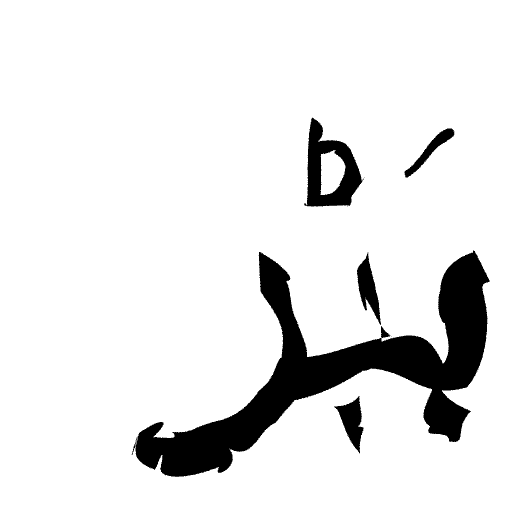} &
  \includegraphics[width=0.085\textwidth]{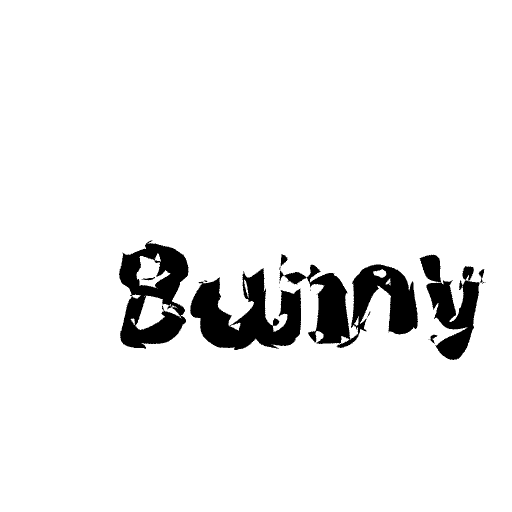} &
  \includegraphics[width=0.085\textwidth]{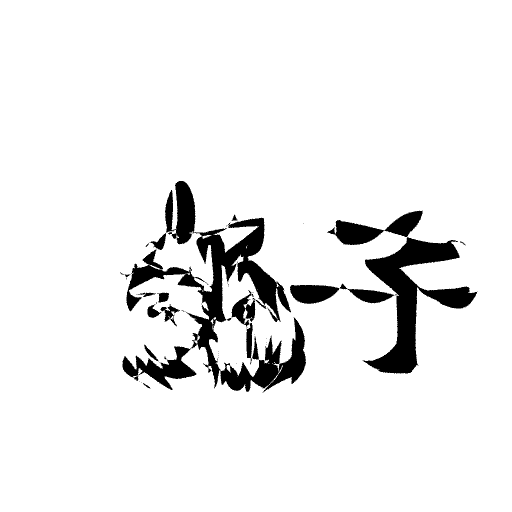} &
  \includegraphics[width=0.085\textwidth]{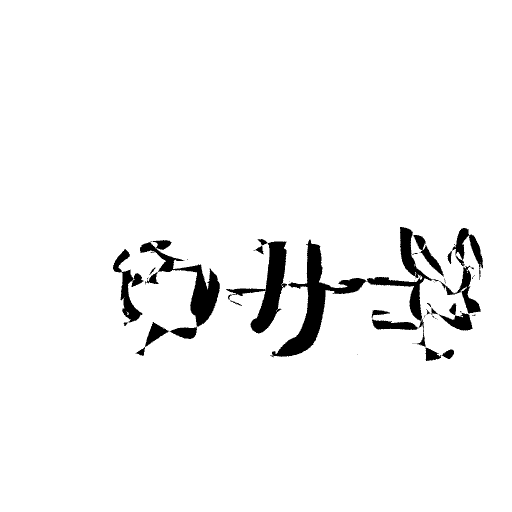} &
  \includegraphics[width=0.085\textwidth]{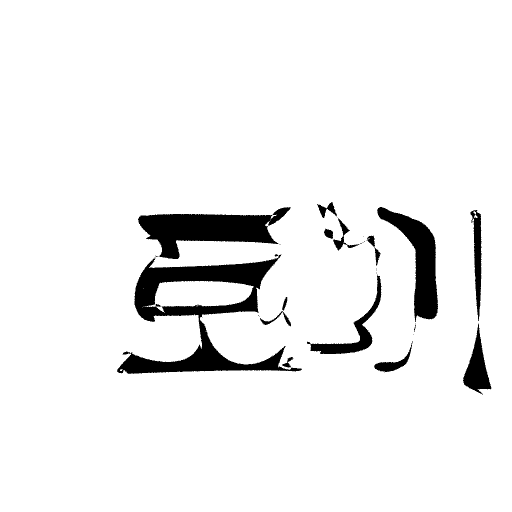} &
  \includegraphics[width=0.085\textwidth]{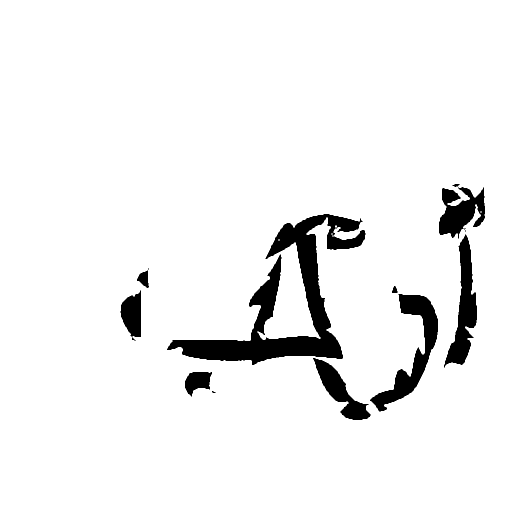} \\

NB &
  \includegraphics[width=0.085\textwidth]{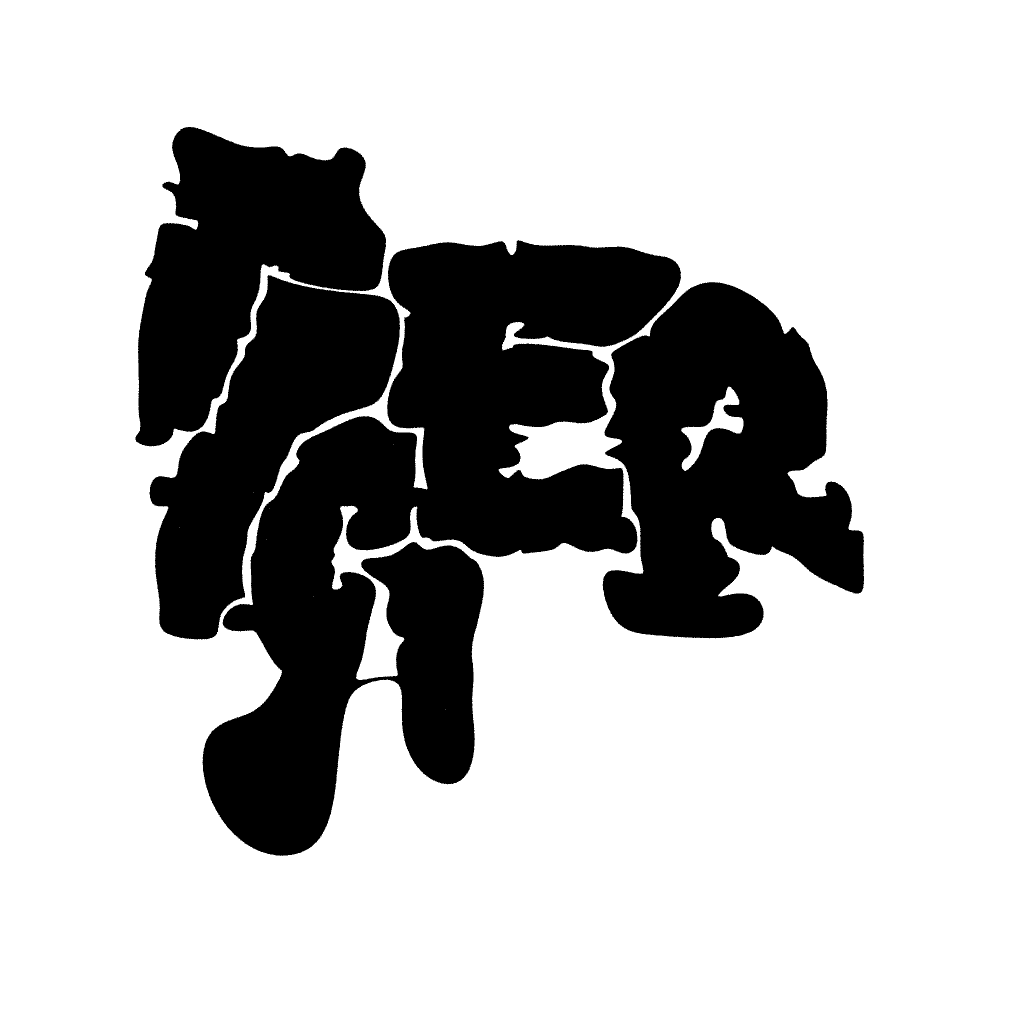} &
  \includegraphics[width=0.085\textwidth]{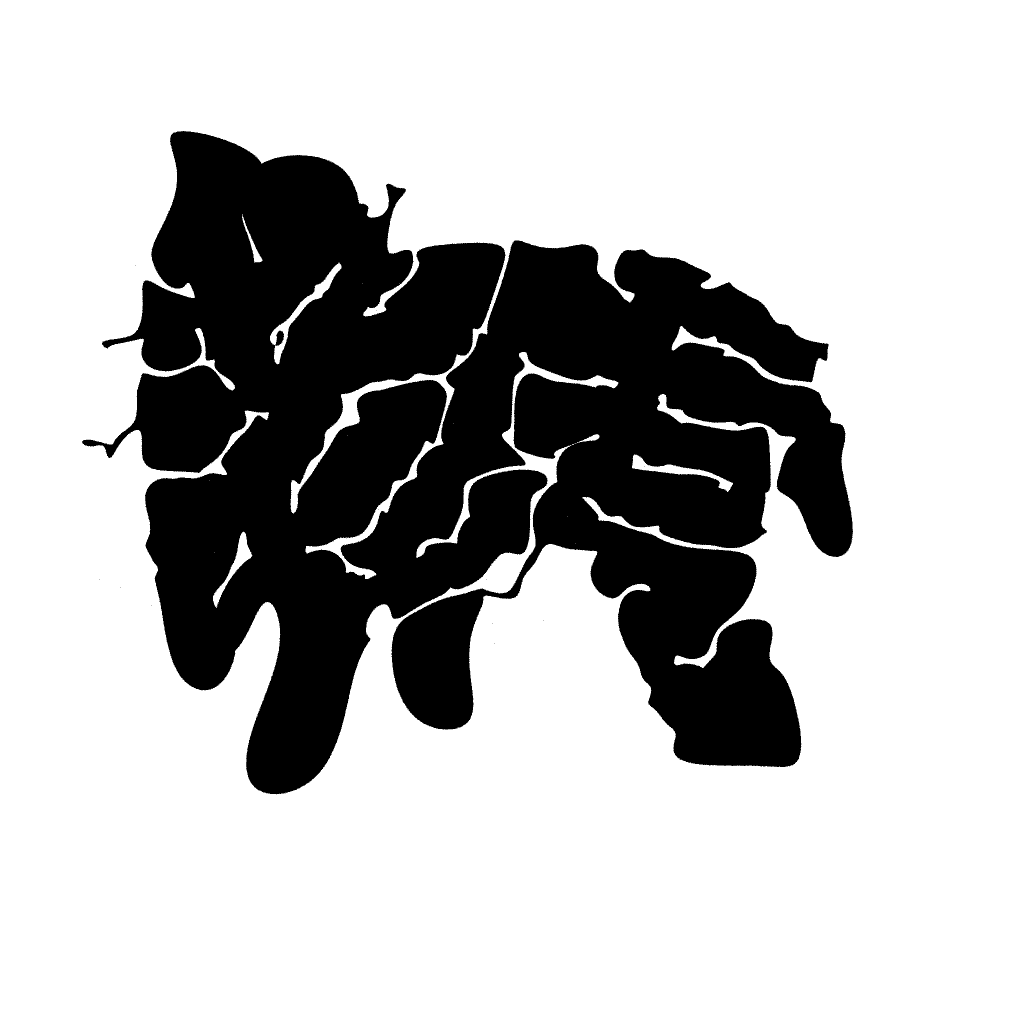} &
  \includegraphics[width=0.085\textwidth]{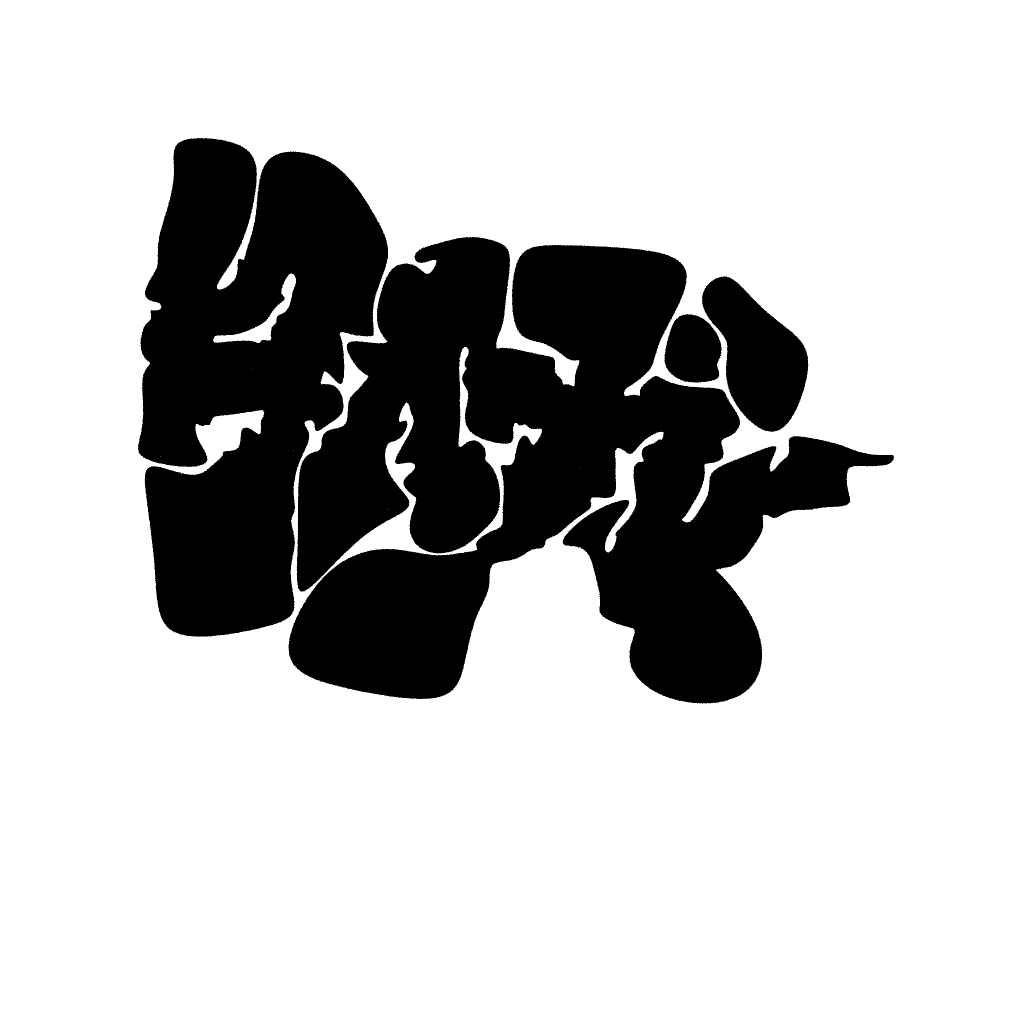} & 
  \includegraphics[width=0.085\textwidth]{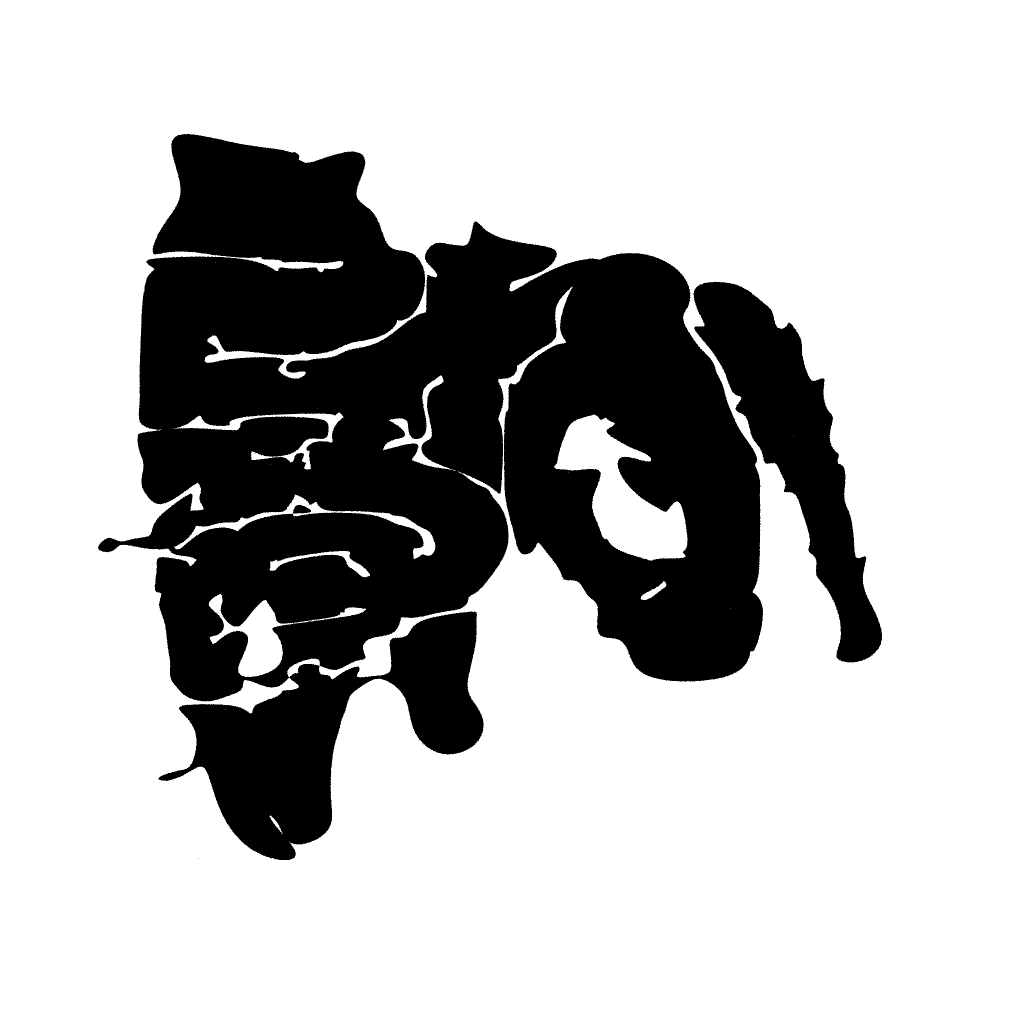} & 
  \includegraphics[width=0.085\textwidth]{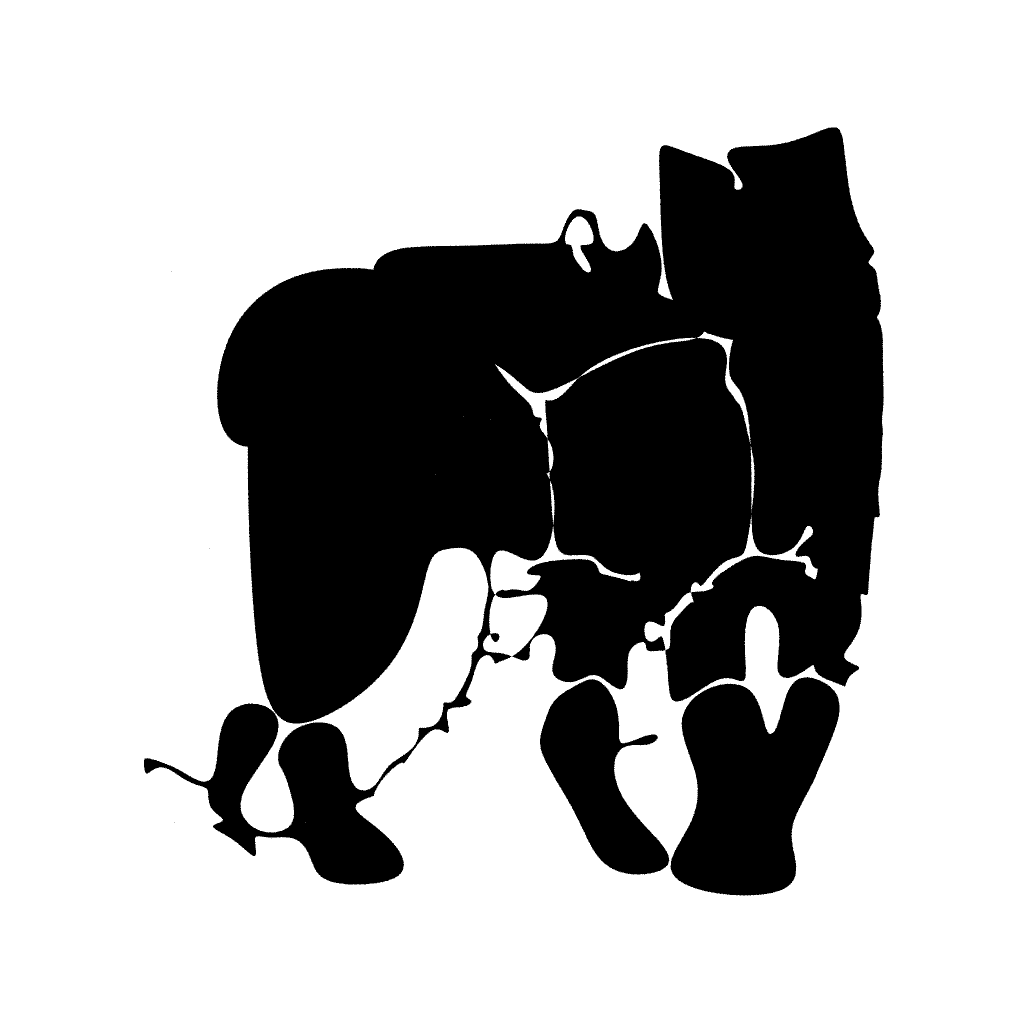} &
  \includegraphics[width=0.085\textwidth]{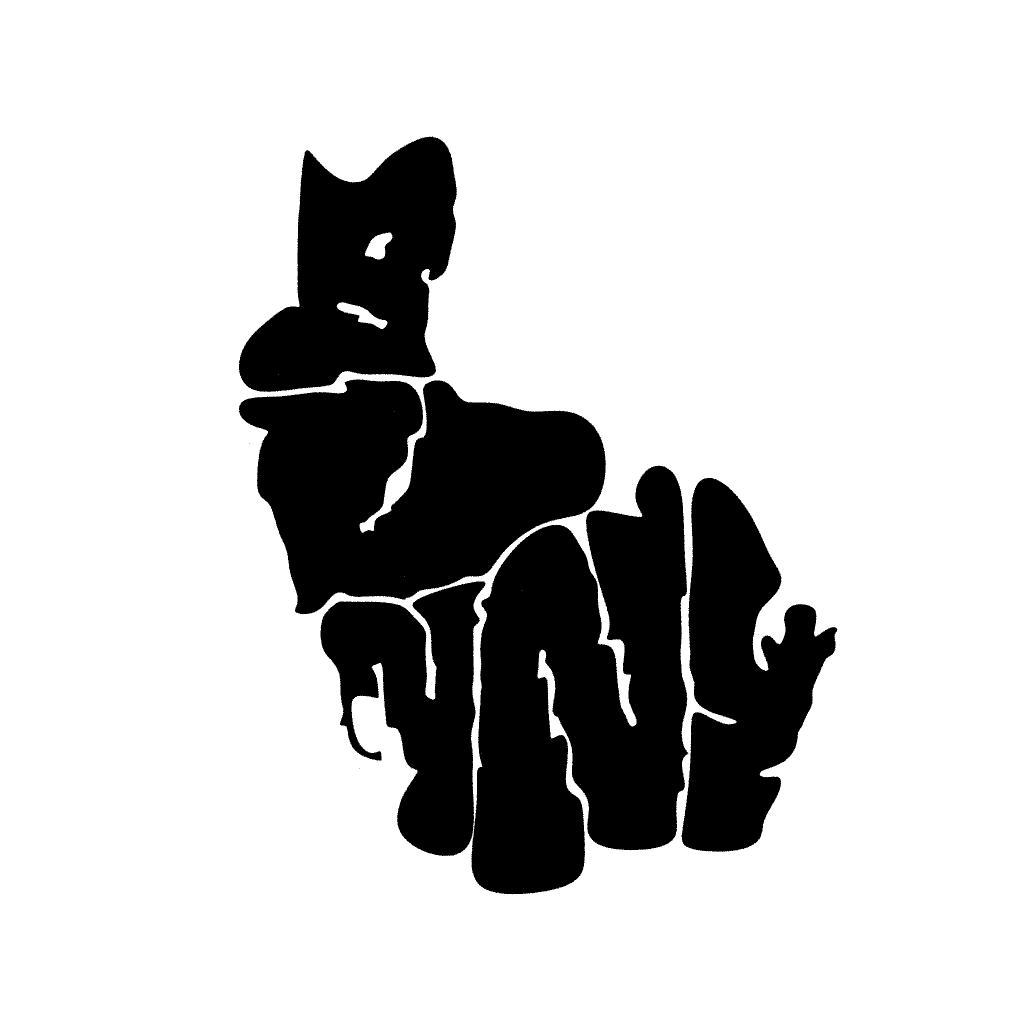} & 
  \includegraphics[width=0.085\textwidth]{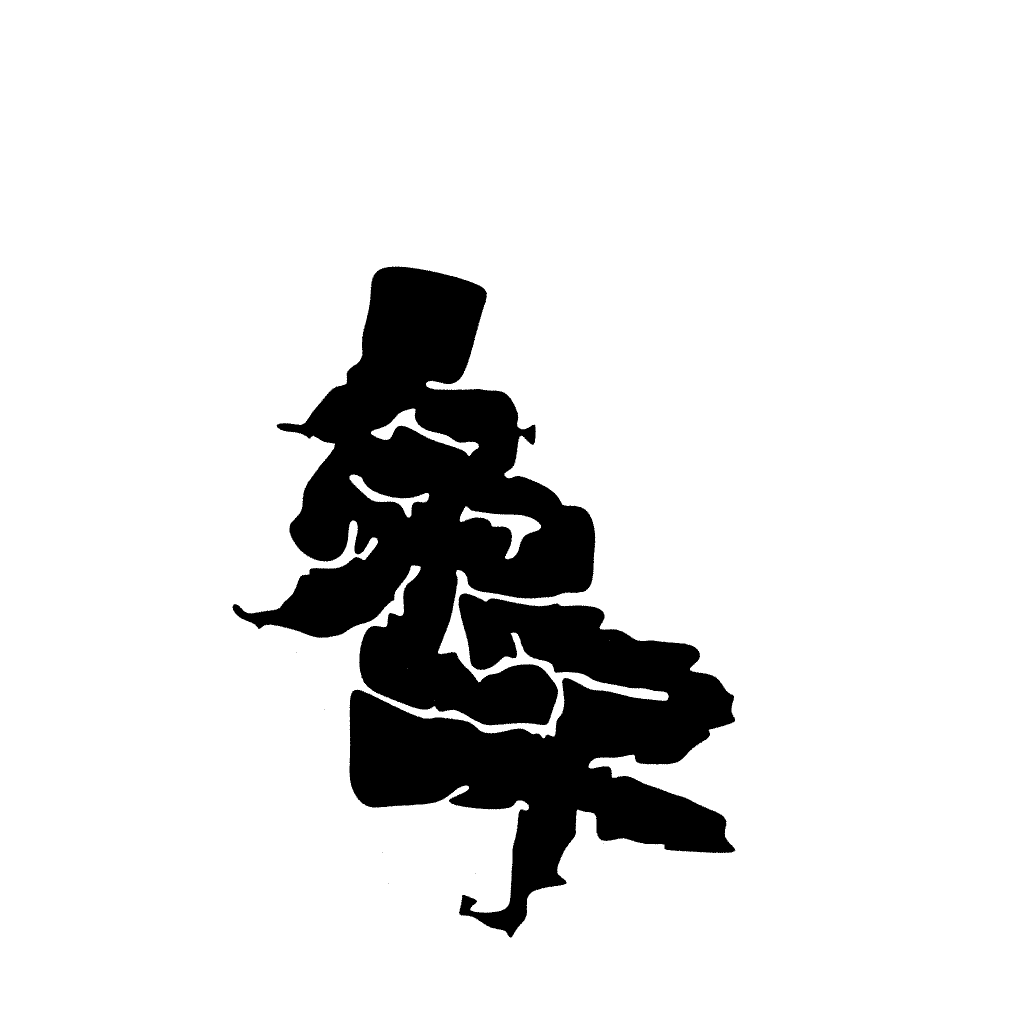} & 
  \includegraphics[width=0.085\textwidth]{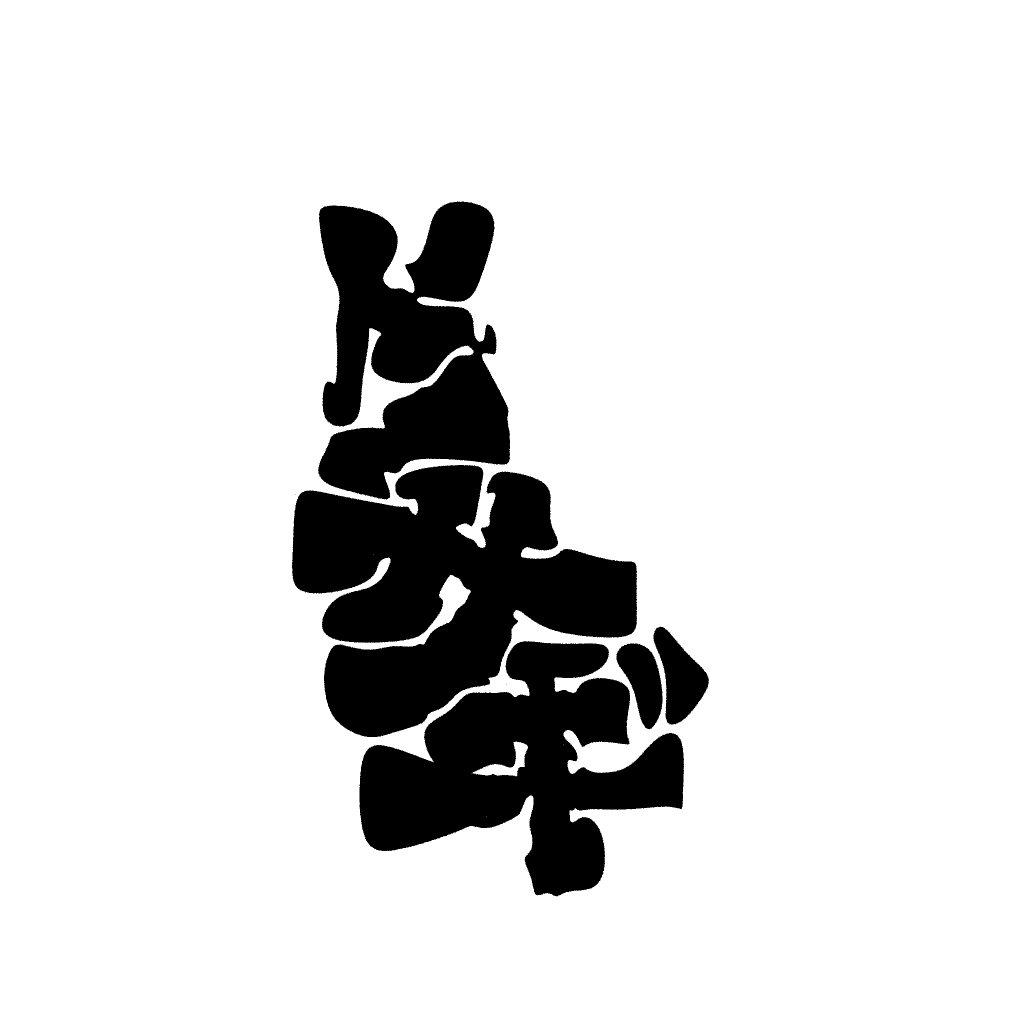} & 
  \includegraphics[width=0.085\textwidth]{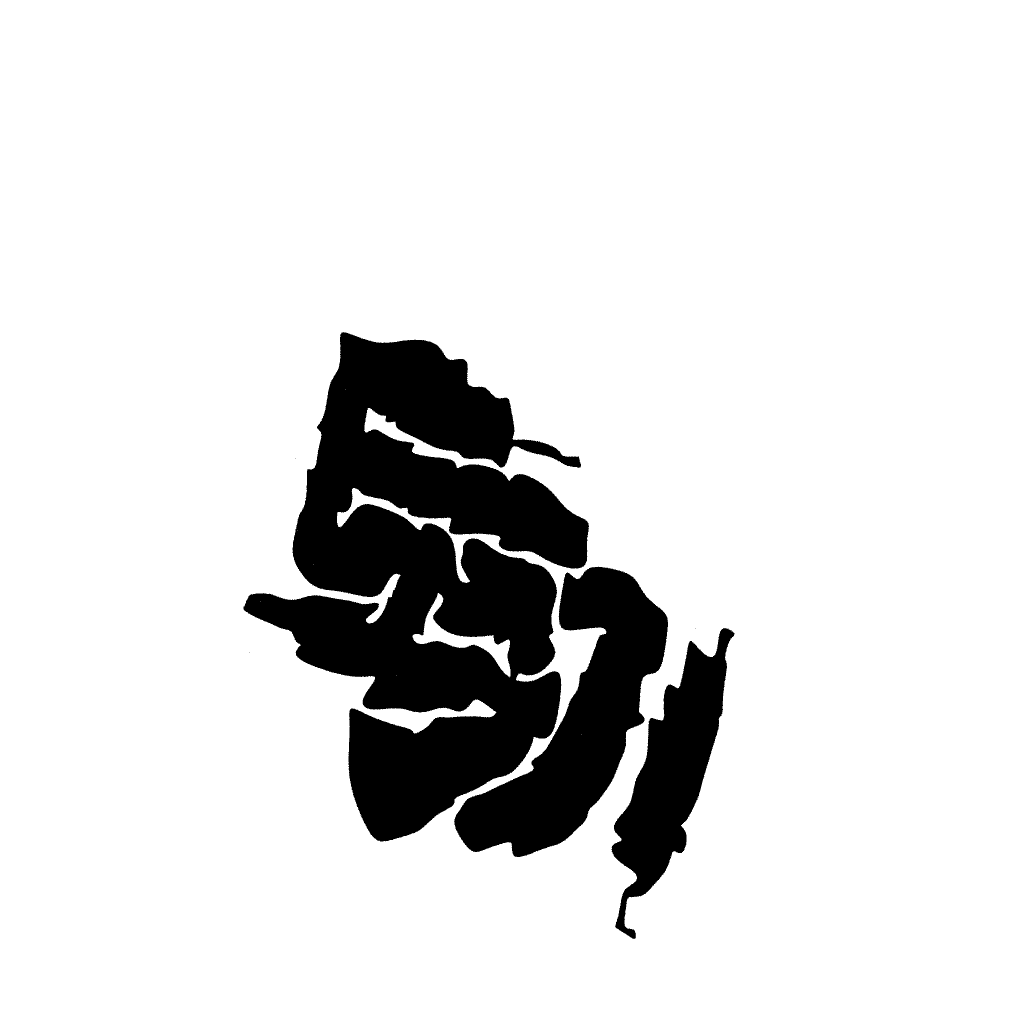} & 
  \includegraphics[width=0.085\textwidth]{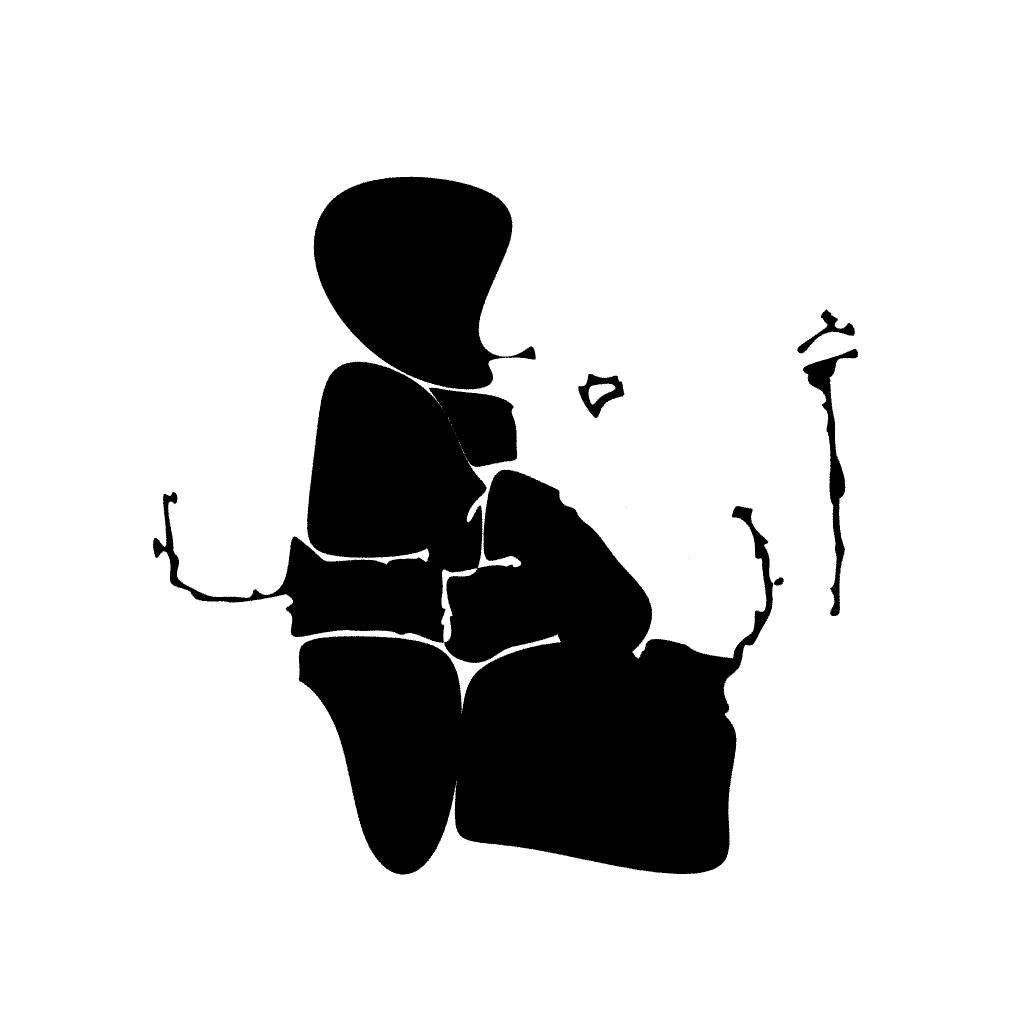} \\

GPT &
  \includegraphics[width=0.085\textwidth]{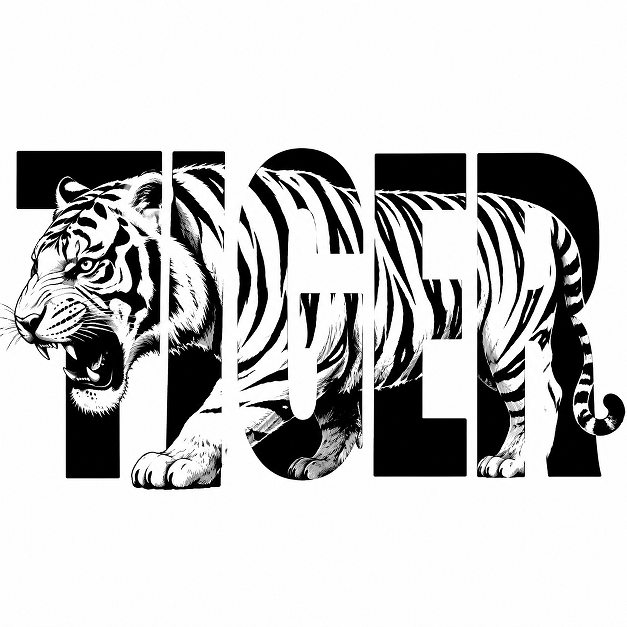} & \includegraphics[width=0.085\textwidth]{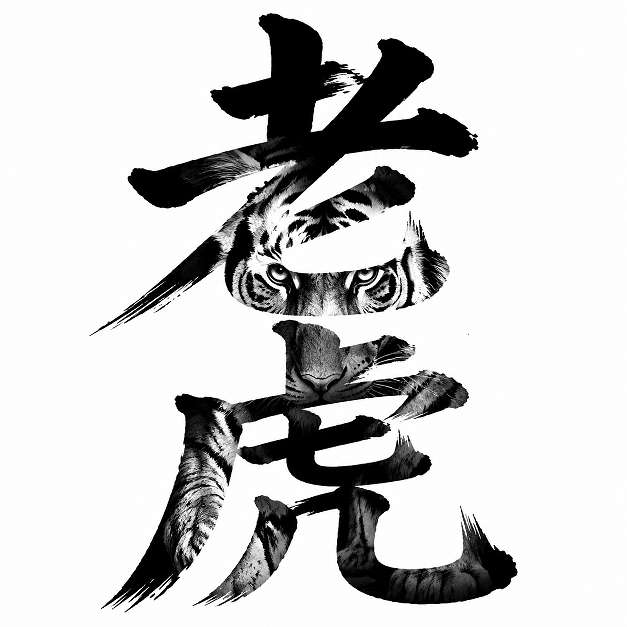} & \includegraphics[width=0.085\textwidth]{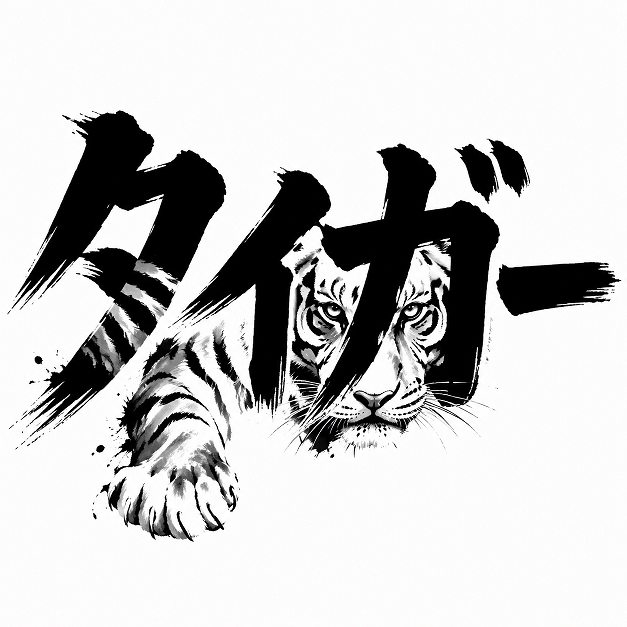} & \includegraphics[width=0.085\textwidth]{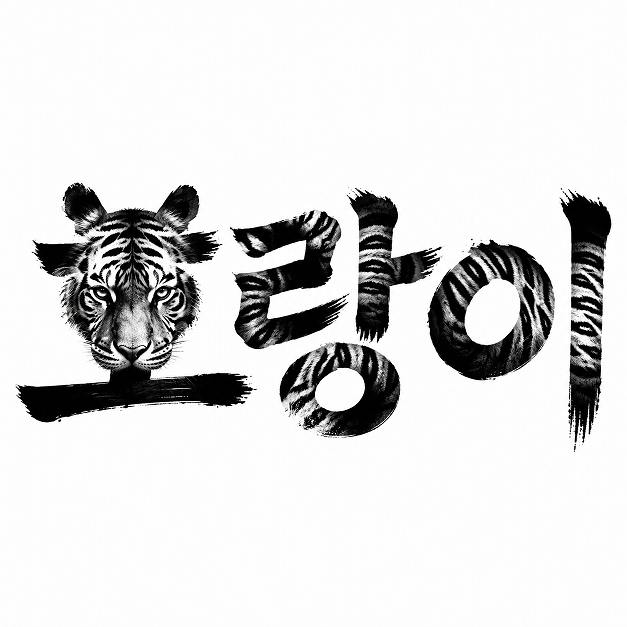} & \includegraphics[width=0.085\textwidth]{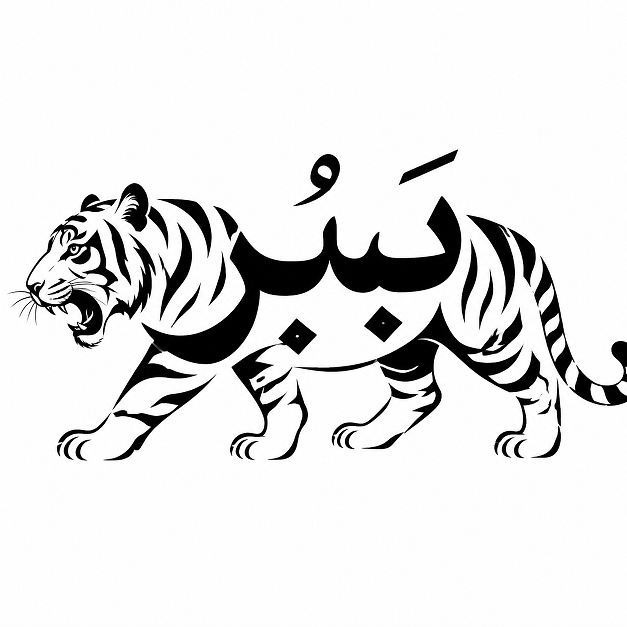} &
  \includegraphics[width=0.085\textwidth]{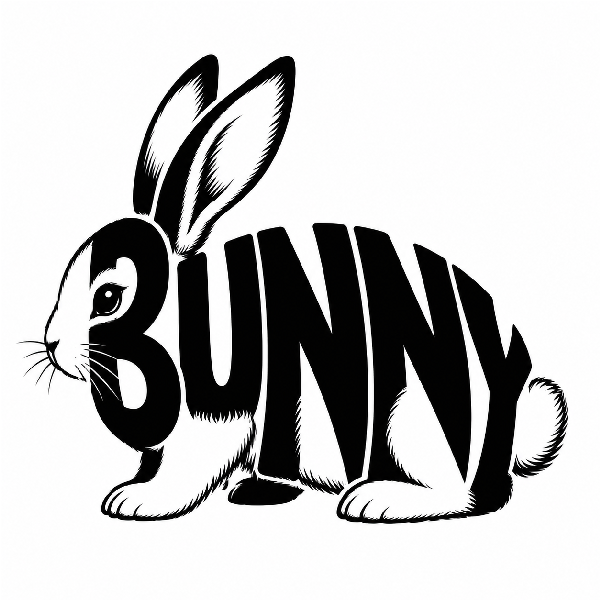} & \includegraphics[width=0.085\textwidth]{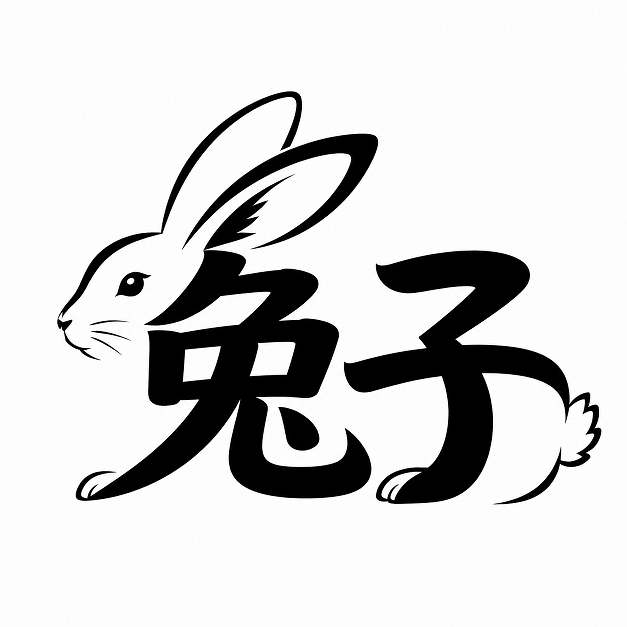} & \includegraphics[width=0.085\textwidth]{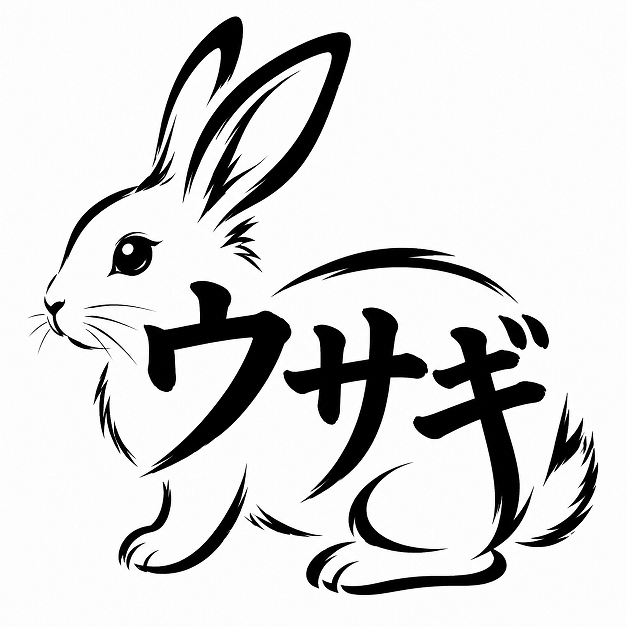} & \includegraphics[width=0.085\textwidth]{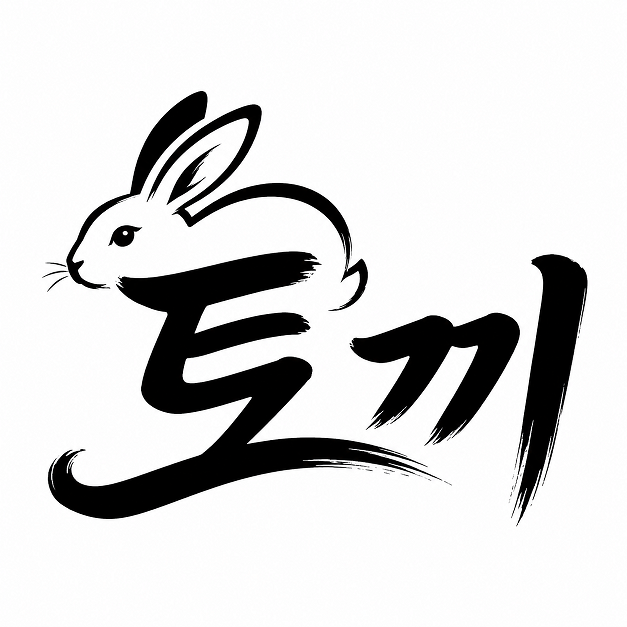} & \includegraphics[width=0.085\textwidth]{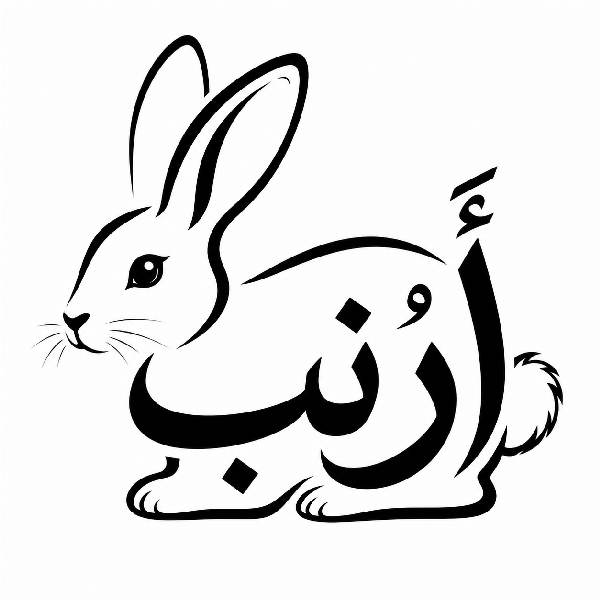} \\
  
Ours &
  \includegraphics[width=0.085\textwidth]{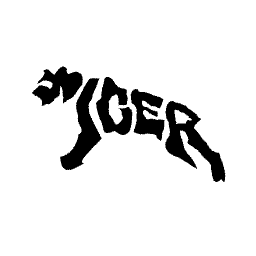} &
  \includegraphics[width=0.085\textwidth]{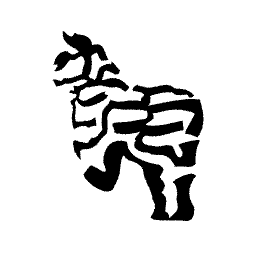} &
  \includegraphics[width=0.085\textwidth]{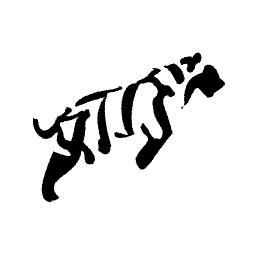} &
  \includegraphics[width=0.085\textwidth]{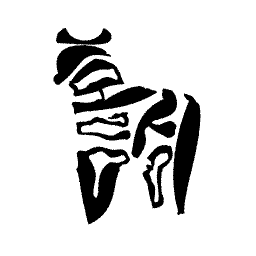} &
  \includegraphics[width=0.085\textwidth]{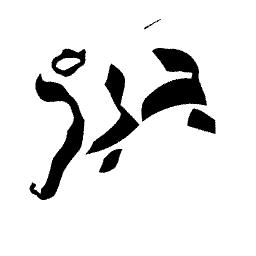} &
  \includegraphics[width=0.085\textwidth]{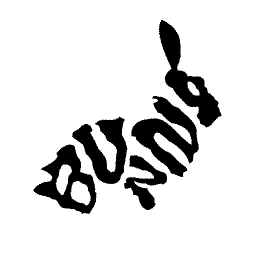} &
  \includegraphics[width=0.085\textwidth]{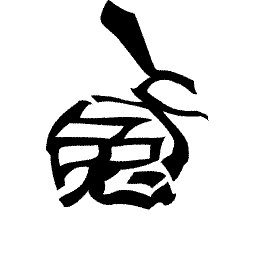} &
  \includegraphics[width=0.085\textwidth]{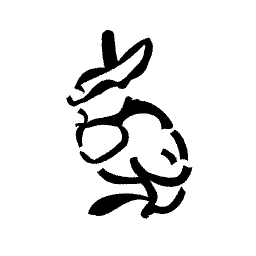} &
  \includegraphics[width=0.085\textwidth]{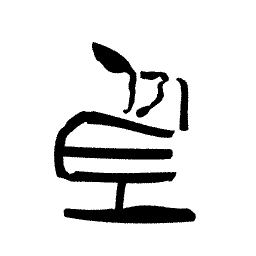} &
  \includegraphics[width=0.085\textwidth]{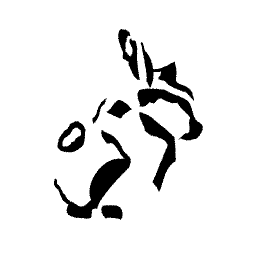} \\

\bottomrule
\end{tabular}

\caption{Comparison with SOTA methods (WAI for Word-As-Image, DT for Dynamic Typography, OBI for OBI-Designer, NB-sp for Neural B‑splines, GPT for GPT-Image-1.5) across 5 languages (English, Chinese, Japanese, Korean, Arabic).}
\label{fig:comparison}
\end{figure*}

\subsection{Local Semantic Deformation}
\label{subsec:local}

The primary objective of the local level is to refine details. In this level, we introduce ControlNet. It can produce more details compared to the filling algorithm employed in the global level. We use the mask image and the text prompt that were utilized for reference optimization in the global level as inputs to ControlNet, and leverage ControlNet to guide Stable Diffusion, compute SDS, and generate the final image. The total loss of the local level is:

\begin{align}
\mathcal{L}_{\text{loc}}=\mathcal{L}_{\text{SDS}}+\lambda_{\text{OCR}}\mathcal{L}_{\text{OCR}}+\lambda_{\text{coll}}\mathcal{L}_{\text{coll}}+\lambda_{\text{ARAP}}\mathcal{L}_{\text{ARAP}}.
\end{align}

\subsubsection{Object recognizability.} Before computing the SDS loss, we apply random data augmentation (including slight scaling, translation, and rotation) to the rendered image to enhance the robustness of optimization, reduce the impact of SDS gradient noise, and improve the semantic consistency of the final graphic generated under spatial transformations. Then, we feed both $I_{mask}$ and the augmented image into ControlNet and Stable Diffusion to compute the SDS loss under the given spatial conditions. The semantic guidance loss $\nabla_{\theta} \mathcal{L}_{\operatorname{SDS}}$ is:

\begin{align}
\mathbb{E}_{t,\epsilon}\left( w(t) \bigl( \hat{\epsilon}(x_t; c_{\text{text}}, c_{\text{ctrl}}, t) - \epsilon \bigr) \frac{\partial \mathcal{R}(\theta)}{\partial \theta} \right),
\end{align}

\begin{align}
\hat{\epsilon} = (1-\gamma)\,\epsilon_{\avr}(x_t; \varnothing, c_{\text{ctrl}}, t) + \gamma\,\epsilon_{\avr}(x_t; c_{\text{text}}, c_{\text{ctrl}}, t).
\end{align}


\subsubsection{Word legibility.} To effectively prevent originally connected or adjacent components of a glyph (e.g., the dot and the main vertical stroke in the letter ``i'') from becoming completely detached or excessively separated, we use the encoder of SuryaOCR~\cite{SuryaOCR} to extract the last layer features of each glyph after the global level. We then compare these features with those of the corresponding individual glyphs in the iteratively updated image, and use the resulting difference as a constraint:

\begin{align}
\mathcal{L}_{\text{OCR}}(\mathbf{X})=\operatorname*{\avr}_{\substack{i}} \Big( \operatorname{MSE} \big(\operatorname{OCR}(X_i), \operatorname{OCR}(X_i^{\text{ref}}) \big) \Big).
\end{align}

Stroke overlap severely compromises character legibility, but neither OCR nor low pass filter can fundamentally eliminate such overlap because they focus only on pixel or feature similarity and lack explicit constraints on spatial separation. Therefore, we introduce a differentiable collision detection based on the signed distance field (SDF)~\cite{SDF, SDFColl} at the geometric level, categorizing collisions into two types: self-intersection within a stroke and inter-stroke collisions. For self-intersection, we apply a differentiable distance computation between Bézier segments, penalizing those whose distance falls below a threshold. The formula is:

\begin{align}
\mathcal{L}_{\text{self}} = \operatorname*{\avr}_{\substack{i\in[1,N]\\ j,k\in C_i}} \left( \operatorname{ReLU}^2 \big( 1-\frac{d_{\min}(\mathbf{q}_{i,j},\mathbf{q}_{i,k})}{\tau} \big) \right),
\end{align}

\noindent where $d_{\min}(\mathbf{q}_{i, j}, \mathbf{q}_{i, k})$ is the differentiable minimum distance based on the SDF ($0$ when intersecting), and $\tau$ is the distance threshold. The calculation of the inter‑stroke collision $\mathcal{L}_{\text{inter}}$ is similar, but operates on curve pairs of different strokes. The overall collision loss is then given by:

\begin{align}
\mathcal{L}_{\text{coll}}=\mathcal{L}_{\text{self}}+\lambda_{\text{inter}}\mathcal{L}_{\text{inter}}.
\end{align}


\begin{table*}[h]
    \centering
    \caption{Quantitative evaluation of CLIP error $\downarrow$ (left) and OCR error $\downarrow$ ($\times 10^{-3}$, right).}
    \label{tab:comparison}
    \small
    \setlength{\tabcolsep}{1pt}
    \begin{subtable}[t]{0.48\textwidth}
        \centering
        \begin{tabular}{l c c c c c}
            \toprule
            Lang. & Ours & WAI & DT & OBI & NB \\
            \midrule
             EN & \textbf{0.737$^{\pm 0.026}$} & 0.754$^{\pm 0.028}$ & 0.773$^{\pm 0.011}$ & 0.757$^{\pm 0.021}$ & 0.741$^{\pm 0.026}$ \\
             ZH & \textbf{0.745$^{\pm 0.023}$} & 0.770$^{\pm 0.024}$ & 0.768$^{\pm 0.008}$ & 0.770$^{\pm 0.016}$ & 0.764$^{\pm 0.020}$ \\
             JA & \textbf{0.743$^{\pm 0.023}$} & 0.759$^{\pm 0.031}$ & 0.769$^{\pm 0.009}$ & 0.764$^{\pm 0.018}$ & 0.768$^{\pm 0.016}$ \\
             KO & \textbf{0.742$^{\pm 0.026}$} & 0.758$^{\pm 0.022}$ & 0.769$^{\pm 0.008}$ & 0.768$^{\pm 0.020}$ & 0.768$^{\pm 0.020}$ \\
             AR & \textbf{0.742$^{\pm 0.027}$} & 0.771$^{\pm 0.014}$ & 0.753$^{\pm 0.016}$ & 0.764$^{\pm 0.017}$ & 0.757$^{\pm 0.021}$ \\
            \midrule
            AVE & \textbf{0.742$^{\pm 0.025}$} & 0.763$^{\pm 0.026}$ & 0.766$^{\pm 0.013}$ & 0.764$^{\pm 0.019}$ & 0.760$^{\pm 0.023}$ \\
            \bottomrule
        \end{tabular}
        \label{tab:clip}
    \end{subtable}
    \hfill
    \begin{subtable}[t]{0.48\textwidth}
        \centering
        \begin{tabular}{l c c c c c}
            \toprule
            Lang. & Ours & WAI & DT & OBI & NB \\
            \midrule
             EN & \textbf{6.097$^{\pm 1.838}$} & 9.607$^{\pm 5.468}$ & 16.05$^{\pm 4.875}$ & 7.086$^{\pm 3.092}$ & 13.41$^{\pm 5.520}$ \\
             ZH & \textbf{6.215$^{\pm 1.734}$} & 6.853$^{\pm 1.994}$ & 7.634$^{\pm 2.374}$ & 6.676$^{\pm 2.448}$ & 8.461$^{\pm 3.145}$ \\
             JA & 5.759$^{\pm 2.242}$ & 4.229$^{\pm 1.579}$ & 5.915$^{\pm 1.988}$ & \textbf{4.179$^{\pm 1.985}$} & 6.017$^{\pm 1.858}$ \\
             KO & 6.923$^{\pm 1.897}$ & 5.744$^{\pm 1.470}$ & 6.453$^{\pm 3.334}$ & \textbf{3.964$^{\pm 1.238}$} & 5.427$^{\pm 1.867}$ \\
             AR & 4.471$^{\pm 1.475}$ & 4.137$^{\pm 1.492}$ & 9.648$^{\pm 2.741}$ & \textbf{3.971$^{\pm 1.978}$} & 5.201$^{\pm 2.683}$ \\
            \midrule
            AVE & 5.893$^{\pm 2.021}$ & 6.119$^{\pm 3.515}$ & 9.139$^{\pm 4.896}$ & \textbf{5.117$^{\pm 2.611}$} & 7.729$^{\pm 4.522}$ \\
            \bottomrule
        \end{tabular}
        \label{tab:ocr}
    \end{subtable}
    
\end{table*}

To maintain the geometric morphology of the final graphic, we construct the Jacobian matrix for each triangular face in the same manner as described above, and further introduce the as-rigid-as-possible (ARAP)~\cite{ARAP} to drive the deformation towards pure rotation, thus achieving local rigidity. Specifically, we obtain the rotation matrix $\mathbf{R}_i$ via polar decomposition of the Jacobian matrix and penalize the norm of the difference between $\mathbf{J}_i$ and $\mathbf{R}_i$. Unlike $\mathcal{L}_{\text{Jac}}$ in the global level, which primarily controls anisotropic scaling and prevents flipping, the ARAP focuses on preserving the rigidity of local shapes. This brings two benefits: first, it maintains morphological stability, preventing misalignment at stroke intersections (e.g., the crossing of the two strokes in ``X'') caused by unconstrained motion of control points; second, it allows strokes to move approximately as a whole. The loss function is:

\begin{align}
\mathcal{L}_{\text{ARAP}}=\operatorname*{\avr}_{\substack{i \in [1, N_f]}} \left ({\left \| \mathbf{J}_i - \mathbf{R}_i \right\|_F^2} \right).
\end{align}


In summary, the local level builds upon the deformation from the global level, introducing semantic guidance and geometric constraints to accomplish the global-to-local optimization. The two levels work synergistically to ultimately generate vector graphics that meet the requirements of both legibility and object recognizability.

\subsection{Implementation Details}
\label{subsec:imp}

The mask image $\mathbf{M}$ can be obtained from Stable Diffusion with region segmentation, hand drawing, or existing images. In our framework, we adopt batch generation using Stable Diffusion XL and segmenting by SAM~\cite{SAM}.

Since the OCR recognition difficulty varies across different language scripts, we independently adjusted the hyperparameters of the OCR loss for each language. In our experiments, the $\lambda_{\text{OCR}}$ is set to 0.2 for Chinese characters and Korean, while set to higher values for other languages (ranging from 0.3 to 0.6). Other hyperparameters are as follows: $\lambda_{\text{ord}}=2.0$, $\lambda_{\text{pix}}=0.002$, $\lambda_{\text{Jac}}=0.02$, $\lambda_{\text{over}}=3000$, $\lambda_{\text{flip}}=100$, $\lambda_{\text{coll}}=1$, $\lambda_{\text{ARAP}}=0.5$, $\lambda_{\text{inter}}=1$.

\section{Experiments and Discussions}
\label{sec:result}


All experiments were conducted on a single NVIDIA RTX 3090 GPU with 24 GB of VRAM. During training, both the global and local levels ran for 500 iterations each, and the total processing time for a single image is less than ten minutes. We selected words with the same semantics in five languages, each of which contains multiple test entries covering 2-5 characters. The input glyphs are based on common fonts: HobeauxRococeaux‑Sherman for English, SimHei for Chinese, Meiryo for Japanese, Malgun Gothic for Korean, and Arial for Arabic. Vector outlines are extracted from these fonts as initial shapes. For each of the 30 test entries (6 words per language across 5 languages), our method generated 15 results per entry. For comparison, Word‑As‑Image~\cite{WordAsImage}, Dynamic Typography~\cite{DynamicTypo}, OBI‑Designer~\cite{OBI}, and Neural B‑splines~\cite{NeuralSplines} produced 10–20 results per entry, while GPT‑Image‑1.5~\cite{openai_gpt_image_1_5} generated one result per entry.

\begin{figure*}[t]
    \centering
    \setlength{\tabcolsep}{2pt}   
     \raisebox{0.35\height}{\rotatebox{90}{\textit{Local  \quad\,\,\,  Global}}}       
    \begin{subfigure}[b]{0.14\linewidth}
        \centering
        \includegraphics[width=\linewidth]{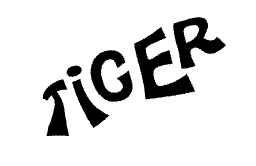}\\
        \includegraphics[width=\linewidth]{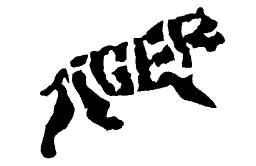}
        \caption{\textbf{Ours}}
        \label{fig:ablation_global_ours}
    \end{subfigure}\hfill
    \begin{subfigure}[b]{0.14\linewidth}
        \centering
        \includegraphics[width=\linewidth]{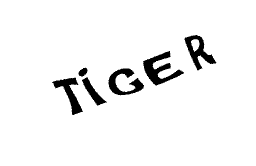}\\
        \includegraphics[width=\linewidth]{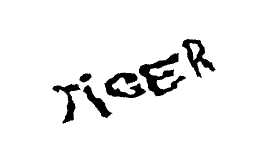}
        \caption{$\mathcal{L}_{\text{fill}}\rightarrow\mathcal{L}_{\text{CLIP}}$}
        \label{fig:ablation_global_fill_clip}
    \end{subfigure}\hfill
    \begin{subfigure}[b]{0.14\linewidth}
        \centering
        \includegraphics[width=\linewidth]{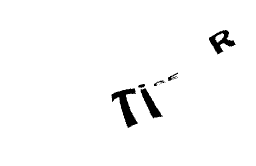}\\
        \includegraphics[width=\linewidth]{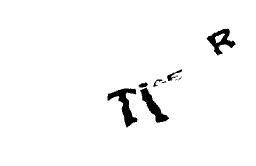}
        \caption{$\mathcal{L}_{\text{fill}}\rightarrow\mathcal{L}_{\text{SDS}}$}
        \label{fig:ablation_global_fill_sds}
    \end{subfigure}\hfill
    \begin{subfigure}[b]{0.14\linewidth}
        \centering
        \includegraphics[width=\linewidth]{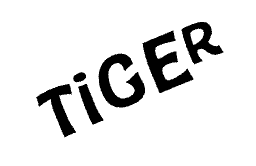}\\
        \includegraphics[width=\linewidth]{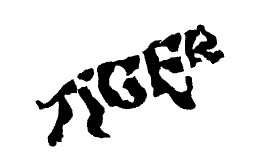}
        \caption{W/o Bézier}
        \label{fig:ablation_global_no_bezier}
    \end{subfigure}\hfill
    \begin{subfigure}[b]{0.14\linewidth}
        \centering
        \includegraphics[width=\linewidth]{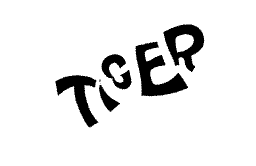}\\
        \includegraphics[width=\linewidth]{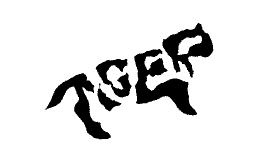}
        \caption{W/o linear}
        \label{fig:ablation_global_no_linear}
    \end{subfigure}\hfill
    \begin{subfigure}[b]{0.14\linewidth}
        \centering
        \includegraphics[width=\linewidth]{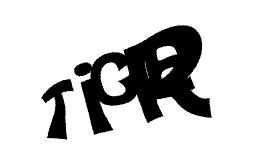}\\
        \includegraphics[width=\linewidth]{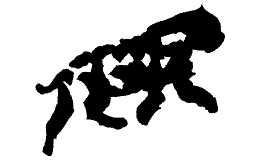}
        \caption{W/o $\mathcal{L}_{\text{ord}}$\&$\mathcal{L}_{\text{pix}}$}
        \label{fig:ablation_global_no_layout}
    \end{subfigure}\hfill
    \begin{subfigure}[b]{0.14\linewidth}
        \centering
        \includegraphics[width=\linewidth]{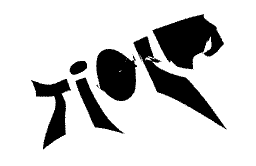}\\
        \includegraphics[width=\linewidth]{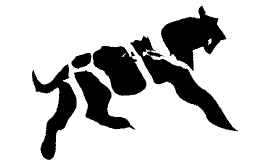}
        \caption{W/o $\mathcal{L}_{\text{Jac}}$}
        \label{fig:ablation_global_no_jac}
    \end{subfigure}

    \medskip   

    \raisebox{0.55\height}{\rotatebox{90}{\,\,\,\,\,\textit{Local}}}  
    \begin{subfigure}[b]{0.14\linewidth}
        \centering
        \includegraphics[width=\linewidth]{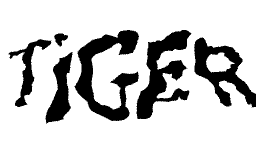}
        \caption{W/o Global}
        \label{fig:ablation_local_no_global}
    \end{subfigure}\hfill
    \begin{subfigure}[b]{0.14\linewidth}
        \centering
        \includegraphics[width=\linewidth]{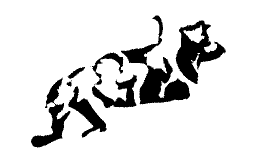}
        \caption{$\mathcal{L}_{\text{SDS}}\rightarrow\mathcal{L}_{\text{CLIP}}$}
        \label{fig:ablation_local_sds_clip}
    \end{subfigure}\hfill
    \begin{subfigure}[b]{0.14\linewidth}
        \centering
        \includegraphics[width=\linewidth]{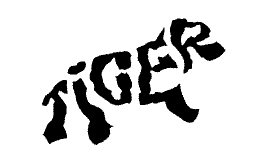}
        \caption{W/o ControlNet}
        \label{fig:ablation_local_no_control}
    \end{subfigure}\hfill
    \begin{subfigure}[b]{0.14\linewidth}
        \centering
        \includegraphics[width=\linewidth]{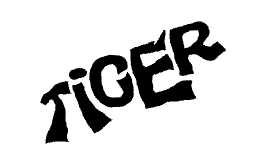}
        \caption{$\mathcal{L}_{\text{SDS}}\rightarrow\mathcal{L}_{\text{fill}}$}
        \label{fig:ablation_local_sds_fill}
    \end{subfigure}\hfill
    \begin{subfigure}[b]{0.14\linewidth}
        \centering
        \includegraphics[width=\linewidth]{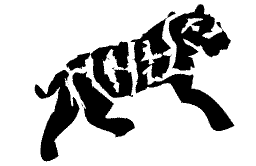}
        \caption{W/o $\mathcal{L}_{\text{ARAP}}$}
        \label{fig:ablation_local_no_jac2}
    \end{subfigure}\hfill
    \begin{subfigure}[b]{0.14\linewidth}
        \centering
        \includegraphics[width=\linewidth]{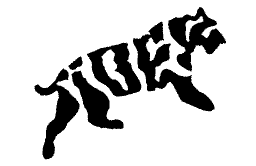}
        \caption{W/o $\mathcal{L}_{\text{OCR}}$}
        \label{fig:ablation_local_no_ocr}
    \end{subfigure}\hfill
    \begin{subfigure}[b]{0.14\linewidth}
        \centering
        \includegraphics[width=\linewidth]{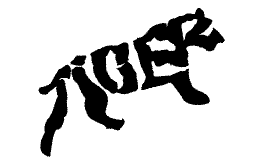}
        \caption{W/o $\mathcal{L}_{\text{coll}}$}
        \label{fig:ablation_local_no_coll}
    \end{subfigure}

    \caption{Ablation study. (a)--(h) Global-level ablation: top row after global deformation, bottom row after full pipeline optimization. (i)--(n) Local-level ablation, showing final results.}
    \label{fig:ablation}
\end{figure*}

\subsection{Qualitative Evaluation}
\label{subsec:qualit}

Fig.~\ref{fig:comparison} shows the experimental results. It can be observed that methods such as Word-As-Image, Dynamic Typography, and OBI-Designer tend to apply only limited deformation when handling multi-characters, or produce results where the original glyphs become unrecognizable after deformation. Neural B‑Spline, on the other hand, primarily fills the target shape with little regard for preserving glyph structure. GPT‑Image-1.5 essentially performs no glyph deformation and instead generates the target image by adding auxiliary forms. In contrast, our method achieves a better balance between object recognizability and word legibility in multi-character scenarios.

\subsection{Quantitative Evaluation}
\label{subsec:quant}

To objectively measure the balance between legibility and recognizability, we adopt two metrics: CLIP error for object recognizability and OCR feature error for word legibility. In our experiments, we observed that severely distorted characters led to extremely low OCR recognition accuracy, with most methods failed to correctly recognize any character. As a result, traditional recognition rate metrics were no longer applicable, and instead we adopted encoding feature differences to measure legibility. CLIP error is computed as the cosine distance between the concave hull region of the generated image and the target text description (e.g., ``a tiger''). The concave hull helps filter out stroke detail interference and focuses on the overall shape. For legibility, we used TrOCR~\cite{li2023trocr} to compute the cosine distance between the encoded features of the deformed characters and those of the initial layout images, and using the initial layout as a baseline reduces positional and layout bias.

Table~\ref{tab:comparison} summarizes the quantitative results. Our method achieves the lowest CLIP distance across all five languages, with an average of~$0.741$, demonstrating consistently superior object recognizability. For OCR feature distance, our method attains the best results on English and Chinese and ranks second on average ($5.869\times 10^{-3}$), only behind OBI‑Designer ($5.003\times 10^{-3}$). The slightly better average OCR score of OBI‑Designer stems from its inherently conservative deformation strategy, which better preserves character identity but sacrifices object recognizability, as evidenced by its substantially higher CLIP distance (averaging~$0.765$). In contrast, our global‑to‑local framework achieves a clearly more favorable trade‑off, significantly improving recognizability while maintaining competitive legibility across all languages.

\subsection{Ablation Study}
\label{subsec:abla}

We perform ablation experiments to examine the contribution of each design component. The results are presented in Fig.~\ref{fig:ablation}, with hyperparameter ablation provided in the supplementary material.

For the global level, we evaluate several variants: replacing the Fill loss with CLIP~(b) or SDS+ControlNet~(c) to test the influence of stochastic noise on layout arrangement; removing the Bézier grid~(d) or the linear transformation~(e) to assess their contribution to deformation capacity; individually ablating the arrangement losses~(f) and the Jacobian loss~(g); and skipping the global stage entirely while directly optimizing glyphs with SDS+ControlNet~(h).

For the local level, we substitute the SDS loss with CLIP~(i), remove ControlNet~(j), or replace with the Fill loss~(k) to verify the role of semantic guidance; remove ARAP~(l) and ablate the OCR loss~(m) or collision loss~(n) to examine their importance for glyph integrity and structural stability.

\subsection{User Study}
\label{subsec:user}

We recruited 22 participants with computer graphics backgrounds to rate 150 results produced by 5 methods across 5 languages and 6 semantics using a 5-point Likert scale~\cite{likert1932technique} (1 = very poor, 5 = excellent) on three criteria: word legibility, object recognizability, and overall quality.

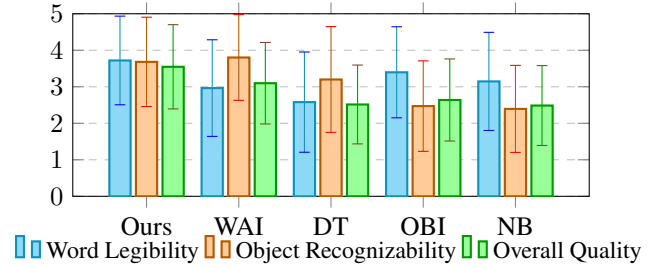
\begin{figure}[!t]
    \centering
    \begin{tikzpicture}
        \begin{axis}[
            width=1\columnwidth,
            height=4cm,
            ybar,
            bar width=8pt,
            ymin=0, ymax=5,
            ytick={0,1,2,3,4,5},          
            symbolic x coords={Ours, WAI, DT, OBI, NB},
            xtick=data,
            enlarge x limits=0.2,
            ymajorgrids=true,
            grid style={dashed, gray!50},
            legend style={
                at={(0.5,-0.18)},          
                anchor=north,
                legend columns=-1,         
                font=\footnotesize,
                draw=none,
                fill=none
            },
        ]

        \addplot+[fill=cyan!40, draw=cyan!70!black, thick]
        plot [error bars/.cd, y dir=both, y explicit]
        coordinates {
            (Ours, 3.721) +- (1.214, 1.214)
            (WAI,  2.964) +- (1.323, 1.323)
            (DT,   2.580) +- (1.373, 1.373)
            (OBI,  3.397) +- (1.247, 1.247)
            (NB,   3.147) +- (1.344, 1.344)
        };

        \addplot+[fill=orange!40, draw=orange!70!black, thick]
        plot [error bars/.cd, y dir=both, y explicit]
        coordinates {
            (Ours, 3.682) +- (1.223, 1.223)
            (WAI,  3.802) +- (1.174, 1.174)
            (DT,   3.200) +- (1.449, 1.449)
            (OBI,  2.471) +- (1.239, 1.239)
            (NB,   2.394) +- (1.192, 1.192)
        };

        \addplot+[fill=green!40, draw=green!60!black, thick]
        plot [error bars/.cd, y dir=both, y explicit]
        coordinates {
            (Ours, 3.547) +- (1.154, 1.154)
            (WAI,  3.098) +- (1.116, 1.116)
            (DT,   2.515) +- (1.080, 1.080)
            (OBI,  2.638) +- (1.124, 1.124)
            (NB,   2.486) +- (1.094, 1.094)
        };

        \legend{Word Legibility, Object Recognizability, Overall Quality}
        \end{axis}
    \end{tikzpicture}
    \caption{User study results: mean scores and standard deviations across five methods and three criteria.}
    \label{fig:user_study}
\end{figure}

Fig.~\ref{fig:user_study} reports the mean scores and standard deviations. Our method achieves the highest ratings on word legibility (3.72) and overall quality (3.55). For object recognizability, it scores 3.68, trailing only Word‑As‑Image (3.80). However, Word‑As‑Image's legibility is much lower (2.96), indicating that it trades off legibility for recognizability. In contrast, our global‑to‑local optimization yields a more favorable balance, as evidenced by the top overall quality score.

\subsection{Limitations and Future Works}
\label{subsec:limitations}
Despite being the first multi-character typography method balancing legibility and recognizability, it remains sensitive to complex masks or simple glyphs, assumes a single connected mask, requires language-specific tuning, and incurs high optimization cost.We plan to extend to multi-component masks via graph-based layout decomposition, reduce mask sensitivity with self-refinement, automate parameter tuning via adaptive learning, accelerate optimization with progressive rendering, and support dynamic/interactive typography.


\section{Conclusion}
\label{sec:conclu}

In this paper, we propose a multi-character semantic typography framework, MSTypography. 
To the best of our knowledge, it is the first semantic typography method tailored for multi-character words. To achieve the best balance between the word legibility and the object recognizability effectively, structural losses and an OCR constraint for character-level readability are designed, while semantic guidance with diffusion priors are introduced.
Instead of local deformation, our method deforms the whole word in a two-level mechanism: it performs mask-driven silhouette approximation at the global level, while semantic-guided refinement at the local level. 
A number of experiments on various languages demonstrate that the proposed method outperforms SOTA methods.

\bibliography{src/references}


\end{document}